\documentclass[manuscript,screen, sigchi]{acmart}
\renewcommand\footnotetextcopyrightpermission[1]{} % removes footnote with conference information in first column
\setcopyright{none}
\AtBeginDocument{%
  \providecommand\BibTeX{{%
    \normalfont B\kern-0.5em{\scshape i\kern-0.25em b}\kern-0.8em\TeX}}}

\usepackage[acronym]{glossaries}
\usepackage{makecell}
\usepackage{lscape}
\usepackage{tabularx}
\usepackage{array}
\usepackage{adjustbox} % text rotate
\usepackage{multirow}
\usepackage{makecell}
\usepackage{caption}
\usepackage{subcaption}
\usepackage{siunitx}                % Load SI unit support. Can be used
\usepackage{tablefootnote}

\definecolor{set1_red}{RGB}{228, 26, 28}
\definecolor{set1_blue}{RGB}{55, 126, 184}
\definecolor{set1_green}{RGB}{77, 175, 74}

\usepackage{textcase} % acronym in headline

\makeglossaries
\newacronym{nnm}{NNM}{Nearest Neighbor Matching}
\newacronym{scm}{SCM}{Sequence Consistency Matching}
\newacronym{nms}{NMS}{non-maximum-suppression}

\newacronym{temporal_ap}{tAP}{Temporal Average Precision}
\newacronym{temporal_iou}{tIoU}{Temporal Intersection over Union}
\newacronym{dataset}{EIGD}{Events in Invasion Games Dataset}
\newacronym{dataset_soccer}{EIGD-S}{}
\newacronym{dataset_handball}{EIGD-H}{}
\newacronym{dataset_provider}{PDD-S}{Private Data-Provider Dataset}
\begin{document}

%\title{Proposing a Unified Taxonomy and Multi-modal Dataset for Events in Invasion Games}
% (Benchmark) + (Multidomain) + (Multimodal) + (Position and Video) + (Invasion Games) + Soccer/Handball 
\title{A Unified Taxonomy and Multimodal Dataset for Events in Invasion Games}


%%%%%%%%%%%%%%%%%%%%%%%%%%%%%%%%%%%% Standard/Expected Format %%%%%%%%%%%%%%%%%%%%%%%%%%%%%%%%%%%%%%%%%%%%%%%%%%%%%%%%%%
\author{Henrik Biermann}
\authornote{Both authors contributed equally to this research.}
\email{h.biermann@dshs-koeln.de}
\orcid{0000-0001-5660-9876}
\affiliation{%
  \institution{Institute of Exercise Training and Sport Informatics, German Sport University Cologne}
  \streetaddress{}
  \city{Cologne}
  \state{}
  \country{Germany}
  \postcode{}
}

\author{Jonas Theiner}
\authornotemark[1]
\orcid{0000-0002-8966-4860}
\email{theiner@l3s.de}
\affiliation{
  \institution{L3S Research Center, Leibniz University Hannover}
  \city{Hannover}
  \state{}
  \country{Germany}
  \postcode{}
}

\author{Manuel Bassek}
\orcid{0000-0002-9394-913X}
\email{m.bassek@dshs-koeln.de}
\affiliation{%
  \institution{Institute of Exercise Training and Sport Informatics, German Sport University Cologne}
  \streetaddress{}
  \city{Cologne}
  \state{}
  \country{Germany}
  \postcode{}
}

\author{Dominik Raabe}
\email{dominik.raabe@dshs-koeln.de}
\orcid{0000-0001-7264-4575}
\affiliation{%
  \institution{Institute of Exercise Training and Sport Informatics, German Sport University Cologne}
  \streetaddress{}
  \city{Cologne}
  \state{}
  \country{Germany}
  \postcode{}
}

\author{Daniel Memmert}
\orcid{0000-0002-3406-9175}
\email{d.memmert@dshs-koeln.de}
\affiliation{%
  \institution{Institute of Exercise Training and Sport Informatics, German Sport University Cologne}
  \streetaddress{}
  \city{Cologne}
  \state{}
  \country{Germany}
  \postcode{}
}

\author{Ralph Ewerth}
\orcid{0000-0003-0918-6297}
\email{ralph.ewerth@tib.eu}
\affiliation{
  \institution{L3S Research Center \\ TIB -- Leibniz Information Centre for Science and
Technology}
  \city{Hannover}
  \state{}
  \country{Germany}
  \postcode{}
}

\renewcommand{\shortauthors}{Biermann and Theiner, et al.}

\begin{abstract}
The automatic detection of events in complex sports games like soccer and handball using positional or video data is of large interest in research and industry.
One requirement is a fundamental understanding of underlying concepts, i.e., events that occur on the pitch.
Previous work often deals only with so-called low-level events based on well-defined rules such as free kicks, free throws, or goals.
High-level events, such as passes, are less frequently approached due to a lack of consistent definitions. This introduces a level of ambiguity that necessities careful validation when regarding event annotations.
Yet, this validation step is usually neglected as the majority of studies adopt annotations from commercial providers on private datasets of unknown quality and focuses on soccer only.
To address these issues, 
we present~(1)~a universal taxonomy that covers a wide range of \textit{low} and \textit{high-level} events for invasion games and is exemplarily refined to soccer and handball,
and~(2)~release two multi-modal datasets comprising video and positional data with gold-standard annotations to foster research in fine-grained and ball-centered event spotting. 
Experiments on human performance demonstrate the robustness of the proposed taxonomy, and that disagreements and ambiguities in the annotation increase with the complexity of the event.
An I3D model for video classification is adopted for event spotting and reveals the potential for benchmarking.
Datasets are available at: \url{https://github.com/mm4spa/eigd}

\end{abstract}

\maketitle

\section{Introduction}\label{sec:intro}

\begin{figure}[bt!]
	\centering
	\includegraphics[width=\linewidth]{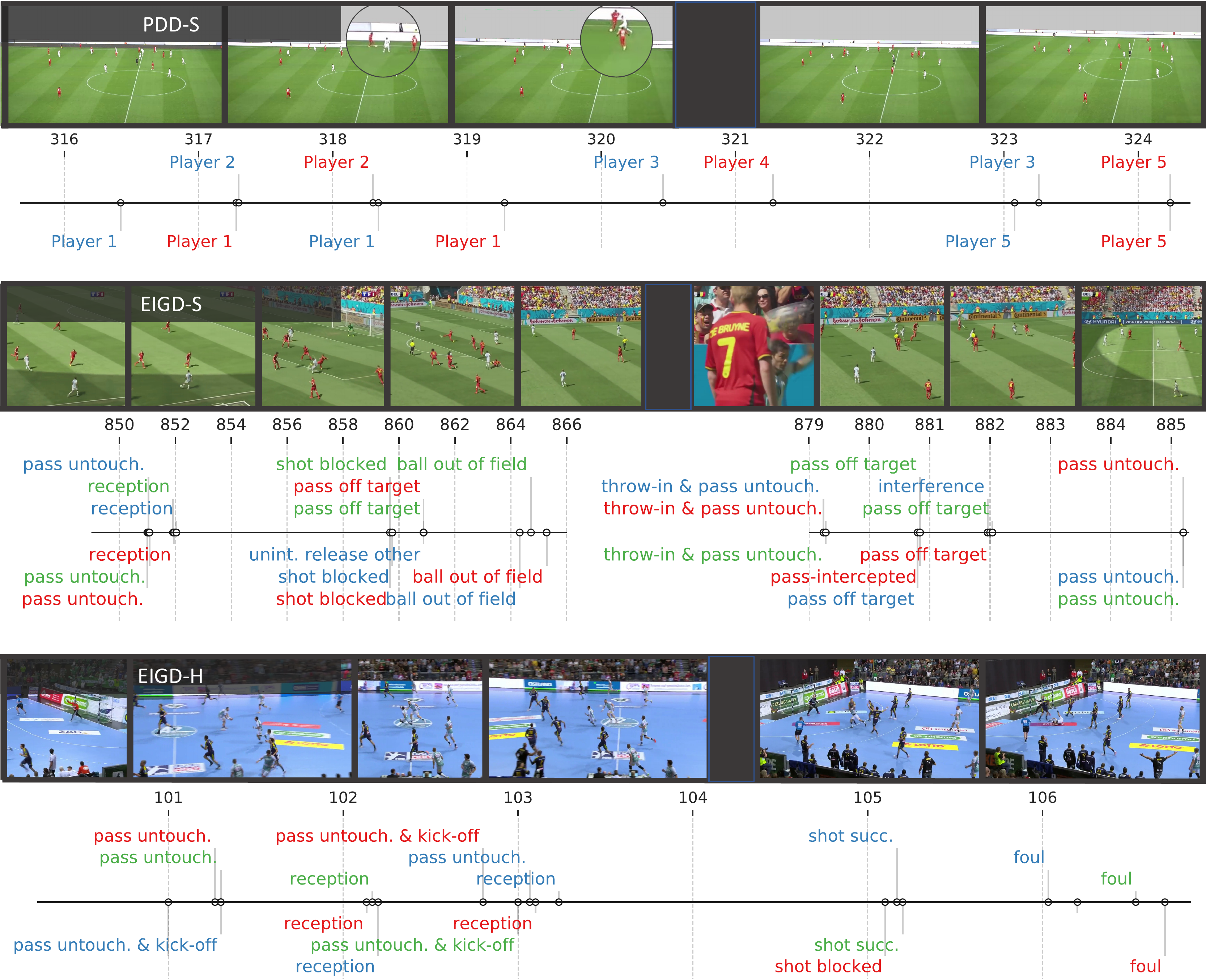}
	\caption{\acrshort{dataset_provider}: Pass annotations from a \textcolor{set1_red}{data provider} vs. an \textcolor{set1_blue}{expert}. \acrshort{dataset_soccer} and \acrshort{dataset_handball}: Example annotations from experienced annotators~(\textcolor{set1_red}{red}, \textcolor{set1_blue}{blue}, \textcolor{set1_green}{green}) using our proposed taxonomy: Despite uncertainties regarding the concrete event type, the annotated timestamp often aligns. The mapping back to shared characteristics such as the motoric skill (e.g., ball release), leads to higher levels of agreement. 
	}
	\label{fig:example_annotations}
\end{figure}

% ----------------------------------- Global intro: motivation, perspectives -----------------------------------

Events play an important role for the (automatic) interpretation of complex invasion games like soccer, handball, hockey, or basketball.
% Three perspectives with respective aims 
Over the last years, three fundamental perspectives emerged with regard to the analysis of sports games, which all value different characteristics of the respective sports: 
(1)~The \emph{sports science} domain demands semantically precise descriptions of the individual developments to analyze success factors~\cite{lamas2014invasion}.
(2)~The \emph{machine learning} community aims to find automatic solutions for specific tasks~(often supervised).
(3)~\emph{Practitioners}~, i.e., coaches or analysts, show little interest in the description of the sport since they are rather interested in the immediate impact of specific modification of training or tactics. 
While the general objective to understand and exploit the underlying concepts in the sports is common to all perspectives, synergistic effects are barely observed~\cite{rein2016big}.
% Capture of events
Descriptive statistics such as possession or shot frequency rely on events that occur on the pitch.
However, collecting semantic and (spatio-)~temporal properties for events during matches is non-trivial, highly dependent on the underlying definitions, and is, in the case of (accurate)~manual annotations, very time-consuming and expensive~\cite{pappalardo2019public}.
% Transition from Events to Automatic annotation of temporal events
While it is a common practice of data providers~\cite{wyscout, opta, stats} for (certain)~matches in professional sport to delegate the annotation of events to human annotators, 
%various works have utilized the increasing amount of data to automate that process.
various approaches have been suggested to automate the process.
% Automatic annotation 
In this respect, the automatic detection of (spatio-)~temporal events has been addressed for (broadcast)~video data~\cite{giancola2018soccernet, giancola2021temporally, sarkar2019generation, vats2020event, sanford2020group, sorano2020automatic, hu2020hfnet, yu2019soccer, jiang2016automatic, liu2017soccer, tomei2021rms, karimi2021soccer, mahaseni2021spotting} and positional data~\cite{sanford2020group, xie2020passvizor, khaustov2020recognizing, chacoma2020modeling, richly2016recognizing, richly2017utilizing, morra2020slicing}.

% -------------------------- Main tasks, boundaries vs spotting, uncertainty, lack of definitions ------------------------
The \textit{temporal event} localization is the task of predicting a semantic label of an event and assigning its start and end time, commonly approached in the domain of video understanding~\cite{caba2015activitynet}.  
Despite a general success in other domains~\cite{lin2019bmn, feichtenhofer2019slowfast, caba2015activitynet, nguyen2018weakly}, it has already been observed that this definition can lead to ambiguous boundaries~\cite{sigurdsson2017actions}.
% Spotting temporal definitions (timestamp)
Sports events can also be characterized by a single representative time stamp~(\emph{event spotting}~\cite{giancola2018soccernet}) and recently there has been success in spotting \textit{low-level} events~\cite{giancola2021temporally, deliege2020soccernet, cioppa2020context} in soccer videos such as goals and cards. 
In contrast, these data acquisition approaches lack more complex, ambiguous, and more frequent events like passes or dribblings that are not covered by existing publicly available~(video) datasets~\cite{feng2020sset, deliege2020soccernet}. Indeed, some definitions of \textit{high-level} events in soccer are provided in the literature~\cite{kim2019attacking, fernandes2019design}, but there is no global annotation scheme or even taxonomy that covers various events that can be evaluated with few meaningful metrics.
Although there are related events in other invasion games such as handball, neither a set of \textit{low-level} and \textit{high-level} events nor a taxonomy are defined in this domain.

% Necessity of large accurate data sets
A shared property for both tasks~(spotting and localization with start and end), regardless of the underlying event complexity, event property (temporal, spatial, or semantic), or data modality~(video or positional data), is the need for labeled event datasets to train and especially to evaluate machine learning approaches.
It is common to integrate~\cite{sanford2020group, fernandez2020soccermap} private event data from data-providers~(e.g., from \cite{wyscout, opta, stats}) of unknown~\cite{liu2013reliability} or moderate~(Figure~\ref{fig:example_annotations}~\acrshort{dataset_provider} as an example) quality.

% -------------------------------------------- Issues in RW summarized ----------------------------------------------
In summary, we observe a lack of a common consensus for the majority of events in the sport.
Neither precise definitions of individual events nor the temporal annotation or evaluation process are consistent. 
Publicly available datasets are uni-modal, focus on soccer, and often consider only a small subset of events that does not reflect the entire match.
These inconsistencies make it for all aforementioned three perspectives difficult to assess the performance of automatic systems and to identify state-of-the-art approaches for the real-world task of fine-grained and ball-centered event spotting from multimodal data sources.

% --------------------------------------- Contributions and Findings ------------------------------------------
% main contribution 1: taxonomy
In this paper, we target the aforementioned problems and present several contributions: 1) We propose a unified taxonomy for \textit{low-level}, and \textit{high-level} ball-centered events in invasion games and exemplary refine it to the specific requirements of soccer and handball. This is practicable as most invasion games involve various shared motoric tasks~(e.g., a ball catch), which are fundamental to describe semantic concepts~(involving intention and context from the game).
Hence, it incorporates various base events relating to \textit{game status}, \textit{ball possession}, \textit{ball release}, and \textit{ball reception}.
% main contribution 2: benchmark datasets
2) We release two multimodal benchmark datasets~(video and audio data for soccer~(\acrshort{dataset_soccer}), synchronized video, audio, and positional data for handball~(\acrshort{dataset_handball})) with gold-standard event annotations for a total of 125 minutes of playing time per dataset.
These datasets contain frame-accurate manual annotations by domain experts performed on the videos based on the proposed taxonomy~(see Figure~\ref{fig:example_annotations}).
In addition, appropriate metrics suitable for both benchmarking and useful interpretation of the results are reported.
% Results and Findings
Experiments on the human performance show the strengths of the \textit{hierarchical} structure, the successful applicability to two invasion games, and reveal the expected performance of automatic models for certain events.
With the increasing complexity of an event~(generally deeper in the \textit{hierarchy}), ambiguous and differing subjective judgments in the annotation process increases.
A case study demonstrates that the annotations from data providers should be reviewed carefully depending on the application.
3) Lastly an \emph{I3D}~\cite{carreira2017quo} model for video chunk classification is adapted for the spotting task using a sliding window and non-maximum suppression and is applied.

The remainder of this paper is organized as follows. 
In Section~\ref{sec:rw}, existing definitions for several events and publicly available datasets are reviewed. The proposed universal taxonomy is presented in Section~\ref{sec:taxonomy}.
Section~\ref{sec:datasets} contains a description of the creation of the datasets along with the definition of evaluation metrics, while Section~\ref{sec:experiments} evaluates the proposed taxonomy, datasets, and baseline concerning annotation quality and uncertainty of specific events. 
Section~\ref{sec:conclusion} concludes the paper and outlines areas of future work.

\section{Related Work}\label{sec:rw}

%To sufficiently cover the wide-ranging research in the scope of our work, 
We discuss related work on events in invasion games~(Section~\ref{rw:event_types}) and review existing datasets~(Section~\ref{rw:datasets}).

\subsection{Events Covered in Various Invasion Games}\label{rw:event_types} % Event Taxonomy

% Dodge et al. Taxonomy of movement patterns
Common movement patterns have been identified in the analysis of spatio-temporal data~\cite{dodge2008towards} such as concurrence or coincidence.
While these concepts are generally applicable to invasion games, % in this work, 
our taxonomy and datasets focus on single actions of individuals~(players), which do not require a complex description of~(team)~movement patterns.
% ----  Handball  ----
For the sport of handball, there are rarely studies on the description of game situations.
However, the influence of commonly understood concepts, such as shots and rebounds has been investigated~\cite{burger2013analysis}.
% ----  Soccer  ----
% Taxonomies vs. Annotation schemes
% Kim et al. offensive taxonomy
In contrast, for soccer, the description of specific game situations has been approached. \citet{kim2019attacking} focus on the attacking process in soccer.
% Fernandes et al. defensive taxonomy
\citet{fernandes2019design} introduce an observational instrument for defensive possessions. The detailed annotation scheme includes 14 criteria with 106 categories %(also considering situational variables, e.g., \textit{game status}, opponent quality, and match location) 
and achieved sufficient agreement in expert studies. However, the obtained semantic description and subjective rating of defensive possessions largely differ from our fine-grained objective approach. 
% Event data + Validation (Pappallardo et al., Liu et al.) 
A common practice for soccer matches in top-flight leagues is to (manually) capture \textit{event data}~\cite{opta, pappalardo2019public}. 
The acquired data describe the on-ball events on the pitch in terms of soccer-specific events with individual attributes. 
While, in general, the inter-annotator agreement for % the modality 
this kind of data has been validated~\cite{liu2013reliability}, especially the \textit{high-level} descriptions of events are prone to errors. 
\citet{deliege2020soccernet} consider 17 well-defined categories which describe meta events, on-ball events, and semantic events during a soccer match. However, due to the focus of understanding a holistic video rather than a played soccer match, only 4 of the 17 event types describe on-ball actions, while more complex events, i.e., passes, are not considered. 
\citet{sanford2020group} spot \textit{passes}, \textit{shots}, and \textit{receptions} in soccer using both positional and video data. However, no information regarding definitions and labels is provided.

\subsection{Datasets}\label{rw:datasets}

To the best of our knowledge, there is no publicly available real-world dataset including positional data, video, and corresponding events, not to mention shared events across several sports.
The majority of datasets for event detection rely on video data and an individual sport domain.
In this context, \emph{SoccerNetV2}~\cite{deliege2020soccernet, giancola2018soccernet} was released, which is a large-scale action spotting dataset for soccer videos. %~(17 classes).
However, the focus is on spotting general and rarely occurring events such as \textit{goals}, \textit{shots}, or cards.
\emph{SoccerDB}~\cite{jiang2020soccerdb} and \emph{SSET}~\cite{feng2020sset} cover a similar set of general events. Even though they relate the events to well-defined soccer rules, they only annotate temporal boundaries.
\citet{pappalardo2019public} present a large event dataset, but it lacks definitions of individual events or any other data such as associated videos.
For basketball, \citet{Ramanathan_2016_CVPR} generated a dataset comprising five types of \textit{shots}, their related outcome (successful), and the \emph{steal event} by using Amazon Mechanical Turk. Here, the annotators were asked to identify the end-point of these events since the definition of the start-point is not clear. 
The \emph{SoccER} dataset~\cite{morra2020soccer} contains synthetically generated data~(positional data, video, and events) from a game engine. 
The volleyball dataset~\cite{ibrahim2016hierarchical} contains short clips with eight group activity labels such as right set or right spike where the center frame of each clip is annotated with per-player actions like standing or blocking. 

To summarize Section~\ref{rw:event_types} and~\ref{rw:datasets}, many studies consider only a subset of relevant~(\textit{low} and \textit{high-level}) events to describe a match.
The quality of both unavailable and available datasets is limited due to missing general definitions~(even spotting vs. duration) apart from well-defined~(per rule) events.

\section{General Taxonomy Design}\label{sec:taxonomy}

\begin{figure*}[tbh]
	\centering
	\includegraphics[width=\textwidth]{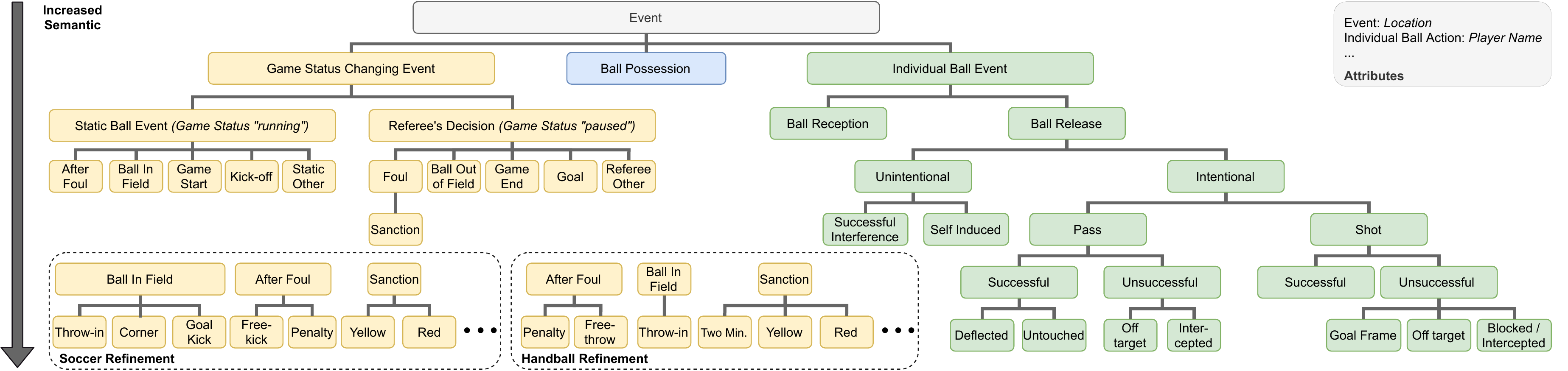}
	\caption{Base taxonomy for invasion games and example refinements for soccer and handball. Starting with basic motoric \emph{individual ball events}, the finer the hierarchy level, the semantic and necessary context information increases.} 
	\label{fig:taxonomy}
\end{figure*}

% Intro sentence
In this section, we construct a unified taxonomy for invasion games that can be refined for individual sports and requirements~(Figure~\ref{fig:taxonomy}). 
Initially, targeted sports and background from a sports science perspective are presented in Section~\ref{subsec:sports}. Preliminaries and requirements for our general taxonomy are listed in Section~\ref{subsec:characteristics}.
Finally, the proposed taxonomy, including concrete event types and design decisions, is addressed in Section~\ref{subsec:categories}.

\subsection{Targeted Sports \& Background}\label{subsec:sports}
% History of family resemblences
Sports games share common characteristics and can be categorized in groups~(\emph{family resemblances})~\cite{Wittgenstein1999}.
Based on that idea,~\cite{Read1997, Hughes2002} structured sports games into three families: (1)~Net and wall games, which are score dependent (e.g., tennis, squash, volleyball), (2)~striking/fielding games, which are innings dependent (e.g., cricket, baseball), and (3)~invasion games, which are time-dependent (e.g., soccer, handball, basketball, rugby). 
In this paper, we focus on the latter. 
Invasion games all share a variation of the same objective: to send or carry an object (e.g., ball, frisbee, puck) to a specific target (e.g., in-goal or basket) and prevent the opposing team from reaching the same goal~\cite{Read1997}. The team that reaches that goal more often in a given time wins. Hence, we argue that the structure of our taxonomy can be applied to all invasion games with a sport-specific refinement of the base events. Please note that we refer to the object in the remainder of this work as a ball for clarity. 

Basic motor skills required in all invasion games involve controlled receiving of, traveling with, and sending of the ball~\cite{Roth2015}, as well as intercepting the ball and challenging the player in possession~\cite{Read1997}. 
Although different invasion games use different handling techniques, they all share the ball as an underlying characteristic. 
Thus, we find that ball events are central for describing invasion games. Moreover, since complex sport-science-specific events such as counterattack, possession play, tactical fouls, or any group activities like pressing are rather sports-specific, we focus on on-ball-ball events in this paper and refer to non-on-ball events as future work.

\subsection{Characteristics \& Unification of Perspective}\label{subsec:characteristics}

We iteratively design the base taxonomy for invasion games to meet certain standards. To provide insights into this process, the following section details the underlying objectives.  

\paragraph{Characteristics}
% Taxonomy attributes 
For the design of a unified taxonomy for invasion games, we view specific characteristics as favorable.
% Hierarchical 
~(1)~A \textit{hierarchical} architecture, in general, is a prerequisite for a clear, holistic structure. We aim to incorporate a format that represents a broad (general) description of events at the highest level and increases in degree of detail when moving downwards in the \textit{hierarchy}. This enables, i.e., an uncomplicated integration of individual annotations with varying degrees of detail as different annotated events (e.g., \textit{shot} and \textit{pass}) can fall back on their common property (here \emph{intentional ball release}) during evaluation.
However, please note that there exists no cross-relation in the degree of detail between different paths~(colors in Figure~\ref{fig:taxonomy}). Events from the same \textit{hierarchical} level may obtain different degrees of detail when from different paths.
% Minimal/non-redundant
~(2)~We target our taxonomy to be \textit{minimal} and \textit{non-redundant} since these characteristics require individual categories to be well-defined and clearly distinguishable from others. In this context, a specific event in the match should not relate to more than one annotation category to support a clear, unambiguous description of the match.
% Exact
~(3)~The taxonomy needs to enable an \emph{exact} description of the match. While the previously discussed \textit{minimal}, \textit{non-redundant} design is generally important, an overly focus on these properties may disallow the description of the \textit{exact} developments in a match. Thus, any neglecting or aggregation for individual categories is carefully considered in the design of the taxonomy.
% Modular expendable
~(4)~Finally, we aim for a \textit{modular expendable} taxonomy. This allows for a detailed examination of specific sports and concepts while still ensuring a globally valid annotation that is comparable (and compatible) with annotations regarding different sports and concepts. 

\paragraph{Unification of Perspectives}
% Representation/Perception of sports (Math vs. Sport science)
The targeted invasion games can generally be perceived from a variety of different perspectives. A mathematical view corresponds to a description of moving objects (players and the ball) with occasional stoppage and object resets (set-pieces reset the ball). On the other hand, a sport-scientist view interprets more complex concepts such as the mechanics of different actions or the semantics of specific situations of play.      
% Previous approaches for unification
To unify these perceptions into a global concept, different approaches such as the \emph{SportsML} language~\cite{SportsML} or \emph{SPADL}~\cite{decroos2019actions} previously targeted a universal description of the match. However, given that the formats originate from a journalist perspective~\cite{SportsML} or provide an integration tool~\cite{decroos2019actions} for data from event providers~(see Section~\ref{exp:data_prov_quality}), they do not pursue the definition of precise annotation guidelines.
% Universal usability 
In contrast, we aim to provide a universal and \textit{hierarchical} base taxonomy that can be utilized by different groups and communities for the targeted invasion games.

\subsection{Annotation Categories}\label{subsec:categories}
% Definition of taxonomy 
The iteratively derived base taxonomy for invasion games is illustrated in Figure~\ref{fig:taxonomy}. Its %individual 
categories and attributes comply with the previously discussed characteristics~(see Section~\ref{subsec:characteristics}) and are outlined in this section~(see Appendix for more detailed definitions).

\subsubsection{Game Status Changing Event}
% First layer
We initialize the first path in our \textit{base taxonomy} such that it corresponds with the most elemental properties of invasion games. Thus, we avoid integrating any semantic information (tactics, mechanics) and regard the so-called, \textit{game status} which follows fixed game rules~\cite{IFAB, IHF}.
% Activeness
The \textit{game status} provides a deterministic division of any point in the match into either active (running) on inactive (paused) play. %(also referred to as the activeness). 
% Changing Events are sufficient
In the sense of a \textit{minimal} taxonomy, we find that an \textit{exact} description of the current \textit{game status} is implicitly included by annotating only those events which cause changes to the current \textit{game status}~(see yellow fields in Figure~\ref{fig:taxonomy}).
% Activeness change events
Moreover, in all targeted invasion games, a shift of the \textit{game status} from active to inactive only occurs along with a rule-based \textit{referee's decision} (foul, ball moving out-of-bounds, game end, or sport-specific stoppage of play) while a shift from inactive to active only occurs along \textit{static-ball-action} (game start, ball in field, after foul, or sport-specific resumption of play). Thus, we discriminate between these two specifications in the path and maintain this \textit{hierarchical} structure.

\subsubsection{Ball Possession}
% Possession
The following paths in our taxonomy comprise additional semantic context to enable a more detailed assessment of individual actions and situations. In this regard, we consider the concept of \textit{possession} (see purple field in Figure~\ref{fig:taxonomy}) as defined by~\citet{link2017individual}. Albeit generally not included in the set of rules of all targeted invasion games (exceptions, i.e., for basketball), the assignment of a team's \textit{possession} is a common practice, and its importance is indicated, for instance, by the large focus of the sports science community~\cite{camerino2012dynamics, casal2017possession, jones2004possession, lago2010game}. 
% Definition
Similar to the \textit{game status}, we only consider the changes to the \textit{possession} with respect to a \textit{minimal} design. 

\subsubsection{Individual Ball Events}

Related to the concept of individual ball \textit{possession} are \textit{individual ball events}, defined as events within the sphere of an individual \textit{possession}~\cite{link2017individual}~(see green fields in Figure~\ref{fig:taxonomy}). 
% Ball control
Along with the definition for an individual \textit{possession}, \citet{link2017individual} define individual \textit{ball control} as a concept requiring a certain amount of motoric skill.
This also involves a specific start and end time for \textit{ball control} which already enables a more precise examination of \textit{individual ball event}.

% Reception
At the start point of individual \textit{possession}, the respective player gains (some degree of) \textit{ball control}. We refer to this moment as a \textit{ball reception}, describing the motoric skill of gaining \textit{ball control}.
% Ball release 
Analogously, at the endpoint of \textit{possession}, the respective player loses \textit{ball control}. We specify this situation as a \textit{ball release}, independent of the related intention or underlying cause. Please note that for a case where a player only takes one touch during an individual \textit{ball control}, we only consider the \textit{ball release} as relevant for the description.
% Dribbling 
For time span between \textit{ball reception} and \textit{ball release}, in general, the semantic concept of a \textit{dribbling} applies. However, various definitions for (different types of) \textit{dribbling} depend on external factors such as the sport, the context, and the perspective. As this semantic ambiguity prevents an \textit{exact} annotation, we do not list \textit{dribbling} as a separate category in the taxonomy but refer this concept to sport-specific refinements.     

% Closing on the physical description
At this point, we utilized the two concepts \textit{game status} and \textit{possession} in invasion games to design the initial \textit{hierarchical} levels for a total of three different paths within the taxonomy~(yellow boxes, purple boxes, and two highest \textit{hierarchical} levels of the green boxes). Accordingly, the required amount of semantic information for the presented levels is within these two concepts. Moreover, since an assessment of these concepts requires low semantic information, we think that the current representation is well-suited for providing annotations in close proximity to the previously presented mathematical perspective on the match. 

% Increased semantic focus
However, we aim for a step-wise approach towards the previously presented sport-scientist perspective for the subsequent \textit{hierarchical} levels.
% Intentional level: P wants to do smth
Therefore, we increase the amount of semantic interpretation by regarding additional concepts, i.e., the overall context within a situation or the intention of players.
% Ball release subcategories 
To this end, we distinguish between two different subcategories for a \textit{ball release}: \textit{intentional} or \textit{unintentional ball release}.
% Involuntary
Regarding an \textit{unintentional ball release}, we generally decide between the categories \textit{successful interference}~(from an opposing player) and \textit{self-induced}~(describing a loss of \textit{ball control} without direct influence of an opponent).
\newpage
% Voluntary
In contrast, for an \textit{intentional ball release}, we further assess the underlying intention or objective of a respective event. We discriminate between a \textit{pass}, including the intention that a teammate receives the released ball, and a \textit{shot} related with an intention (drawn) towards the target. 
% Interference
In some rare cases, the assessment of this intention may be subjective and difficult to determine. However, specific rules in invasion games require such assessment, i.e., in soccer, the goalkeeper is not allowed to pick up a released ball from a teammate when it is "\emph{deliberately}" kicked towards him~\cite{IFAB}.

% Double annotation for set-pieces
Please note that we define \textit{individual ball events} as mutually exclusive, i.e., only one event from that path can occur at a specific timestamp. However, a single point in time may generally include multiple events from different paths in the \textit{taxonomy}. 
Examples for this, in particular, are set-pieces. Here, we annotated a \textit{static-ball event} such as \emph{ball in field}, indicating the previously discussed shift of the \textit{game status}, and an \textit{individual ball event} (e.g., a \textit{pass}) describing the concrete execution. This is necessary as the type of \textit{static-ball event} does not definitely determine the type of \textit{individual ball event}~(i.e., a free-throw in basketball can theoretically be played as a pass from the rim). Nevertheless, since each set piece involves some sort of \textit{ball release}~(per definition of rule, a set-piece permits a double contact of the executing player), an \textit{exact} annotation of set-pieces is provided by the implicit link of simultaneous~(or neighboring) \textit{ball release} and \textit{static ball events}.

\subsubsection{Attributes}
% Attributes in general
A global method to add semantic information to the annotation is provided by defining specific \textit{attributes} for certain events. 
% Form of Annotation
While not representing a specific path in the \textit{base taxonomy}, an \textit{attribute} is defined as a name or description like \emph{pixel location} of the event in the video~(Figure~\ref{fig:taxonomy} upper-right) and thus provides additional information to the respective event.
% Position and implications
When an \textit{attribute} is defined for an event at a certain \textit{hierarchical} level, it is valid for all child events in lower levels.

\section{Events in Invasion Games Dataset}\label{sec:datasets}

\begin{table}[b]
\caption{Dataset distribution: Approx. 40\,\% of all events are reserved for testing, i.e., two of five matches, respectively.}
\label{tab:stats}
\centering
\small
\fontsize{7}{10}\selectfont
\def\arraystretch{0.7}
%%%%%

\begin{tabularx}{\linewidth}{Xlrr}
\toprule
Shared Parent Event & Event &    \acrshort{dataset_soccer} &  \acrshort{dataset_handball} \\
\midrule \midrule
\textbf{ball possession change} &  &            171 &              136 \\ \hline
\textbf{ball reception} &  &           923 &            2268 \\ \hline
\textbf{ball release} &  &           1531 &            2470 \\ \hline
\textbf{pass} &  &            1346 &              2292 \\
 & intercepted &            83 &              14 \\
                           & off target &           175 &               9 \\
                           & successful deflected &            24 &               6 \\
                           & successful untouched &          1064 &            2263 \\
\textbf{shot} &  &            31 &              175 \\
                           & blocked/intercepted &   17 &              61 \\
                           & goal frame &             0 &               8 \\
                           & off target &             8 &              12 \\
                           & successful &             6 &              94 \\
\textbf{unintentional} & other &            74 &               0 \\
                           & successful interference &            80 &               3 \\ \hline
\textbf{referee decision}         &        &            142 &               252 \\ \hline
                           & ball out of field &           101 &              21 \\
                           & foul &            32 &             114 \\
                           & goal &             3 &              86 \\
                           & other &             5 &              10 \\
                           & two min &             n.d. &               8 \\
                           & yellow &             1 &              13 \\ \hline
\textbf{static ball action}         &         &            121 &               207 \\ \hline
                           & corner &            11 &               n.d. \\
                           & free-kick &            29 &              84 \\
                           & game start &             1 &               2 \\
                           & goal-kick &            18 &               n.d. \\
                           & kick-off &             1 &              87 \\
                           & other &             0 &               6 \\
                           & penalty &             1 &              11 \\
                           & throw-in &            60 &              17 \\
\bottomrule
\end{tabularx}

%%%%%

\end{table}

The following Section~\ref{sec:dataset_description} describes our multimodal (video, audio, and positional data) and multi-domain (handball and soccer) dataset for ball-centered event spotting~(\acrshort{dataset}). 
In Section~\ref{sec:metrics}, appropriate metrics for benchmarking are introduced.

\subsection{Data Source \& Description}\label{sec:dataset_description}
To allow a large amount of data diversity as well as a complete description of a match, we regard longer sequences from different matches and stadiums. 
We select 5 sequences à 5 minutes from 5 matches resulting in 125 minutes of raw data, respectively, for handball and soccer.

\paragraph{Data Source}
For the handball subset, referred as \acrshort{dataset_handball}, synchronized video and positional data from the first German league from 2019 are kindly provided by the Deutsche Handball Liga and Kinexon\footnote{\url{https://kinexon.com/}} and contain HD~($1280 \times 720$ pixels) videos at 30\,fps and positional data for all players in 20\,Hz.
The videos include unedited recordings of the game from the main camera~(i.e., no replays, close-ups, overlays, etc.).
Some events are more easily identified from positional data, other events that require visual features can be extracted from video, making \acrshort{dataset_handball} interesting for multimodal event detection.

For the soccer dataset, referred to as \acrshort{dataset_soccer}, we collect several publicly available broadcast recordings of matches from the FIFA World Cup (2014, 2018) in 25\,fps due to licensing limitations without positional data.
A characteristic for annotation purpose is naturally given regarding the utilized videos. \acrshort{dataset_soccer} includes difficulties for the annotation, such as varying camera angles, long replays, or the fact that not all players are always visible in the current angle, making it challenging to capture all events for a longer sequence.
% Consequences
All events are included that are either visible or can be directly inferred by contextual information~(e.g., timestamp of a \emph{ball release} is not visible due to a cut from close-up to the main camera). However, with exceptions for \emph{replays} as these do not reflect the actual time of the game.

\paragraph{Annotation Process \& Dataset Properties}\label{sec:dataset:annotation_process}

To obtain the dataset annotations, the general task is to spot the events in the lowest \textit{hierarchy} level since the parent events~(from higher \textit{hierarchy} levels) are implicitly annotated. Therefore, the taxonomy~(Section~\ref{sec:taxonomy}) and a concrete annotation guideline including definitions for each event, examples, and general hints~(see Appendix) were used.
An example for \emph{unintentional ball release - self-induced} is this situation: A player releases the ball without a directly involved opponent, e.g., slips/stumble or has no reaction time for a controlled \textit{ball release}. For instance, after an \textit{intercepted/blocked pass} or \textit{shot} event. Timestamp: on \textit{ball release}.

We hired nine annotators~(sports scientists, sports students, video analysts). 
Due to the complexity of soccer, four of them annotated each sequence of the \acrshort{dataset_soccer} test set.
Note, that two of the five matches are indicated as a test set~(\acrshort{dataset_soccer}-T, \acrshort{dataset_handball}-T), respectively.
In addition, one inexperienced person without a background in soccer annotated the \acrshort{dataset_soccer}-T.
For \acrshort{dataset_handball}, three experienced annotators also processed each sequence of the test set.
An experienced annotator has labeled the remaining data~(e.g., reserved for training).
The annotation time for a single video clip is about 30 minutes for both datasets.
The number of events given the entire dataset and one expert annotation is presented in Table~\ref{tab:stats} and we refer to Section~\ref{exp:aggreement} for the assessment of the annotation quality.
Figure~\ref{fig:example_annotations} shows two sequences with annotations from several persons as an example.

For each dataset, we assess the human performance~(see Section~\ref{exp:aggreement}) for each individual annotator. The annotation with the highest results is chosen for release and as reference in the evaluation of the baseline~(see Section~\ref{exp:baseline}).

\subsection{Metrics}\label{sec:metrics}

% Events with a duration
For events with a duration, it is generally accepted~\cite{lin2019bmn, caba2015activitynet} to measure the \acrfull{temporal_iou}. 
This metric computes for a pair of annotations by dividing the intersection~(overlap), by the union of the annotations. The \acrshort{temporal_iou} for multiple annotations is given by the ratio of aggregated intersection and union. 
% Events with timestamp
% Tolerance 
Regarding events with a fixed timestamp, a comparison between annotations is introduced in terms of temporal tolerance. Thereupon, when given a predicted event, count a potentially corresponding ground-truth event within the respective class as a true-positive, if and only if it falls within a tolerance area~(in seconds or frames). 
% Problem of corresponding events
Yet, the definition of corresponding events from two different annotations is non-trivial~\cite{sanford2020group, deliege2020soccernet, giancola2018soccernet}, especially for annotations with different numbers of annotated events. 
% Nearest Neighbour
A common method to circumvent this task is to introduce a simplification step using a \acrfull{nnm} which considers events with the same description on the respective \textit{hierarchy} level. 
%to allow a computation of metrics like precision and recall for a fixed temporal tolerance area~\cite{sanford2020group}. 
After defining true positive, false positive, and false negative, this enables the computation of the \acrfull{temporal_ap}~(given by the average over multiple temporal tolerance areas~\cite{deliege2020soccernet, giancola2018soccernet}) or precision and recall for a fixed temporal tolerance area~\cite{sanford2020group}.

% Nearest Neighbour Bias
However, as the \acrshort{nnm} generally allows many-to-one mappings, a positive bias is associated with it. For instance, when multiple events from a prediction are assigned to the same ground-truth event~(e.g., \textit{shot}), they might all be counted as true positives~(if within the tolerance area), whereas the mismatch in the number of events is not further punished. This bias is particularly problematic for automatic solutions that rely on~(unbiased) objectives for training and evaluation. 
% Non Maximum Suppression
Therefore, \citet{sanford2020group} apply a \acrfull{nms} which only allows for a single prediction within a respective \acrshort{nms} window. While this presents first step, a decision on the (hyper-parameter) \acrshort{nms} window length can be problematic. When chosen too large, the \acrshort{nms} does not allow for a correct prediction of temporally close events. In contrast, when chosen too small, the \acrshort{nms} only partially accounts for the issue at hand. Moreover, the lack of objectivity draws a hyper-parameter tuning, e.g., a grid search, towards favoring smaller window lengths for \acrshort{nms}. 

% Contribution: SCM 
To avoid these issues, we propose an \emph{additional} method to establish a one-to-one mapping for corresponding events from two annotations~(with possibly different numbers of events). 
% Equal number of events in practice
In theory, this mapping can only be established if the number of event types between the annotations is equal. However, in practice, this requirement is rarely fulfilled for the whole match. Moreover, even when fulfilled, possibly additional and missing events might cancel each other out. 
% Sequence consistency definition
Based on this, a division of the match into independent~(comparable) segments is a reasonable pre-processing step. Thus, we define a \textit{sequence} as the time of an active match between two \textit{game status-changing events}~(objectively determined by the set of rules~\cite{IFAB, IHF}). Then, (i)~we count the number of \textit{sequences} in two \textit{annotations} to verify that no \textit{game status changing events} were missed~(and adopt \textit{game status changing events} that were missed), (ii)~count the number of annotated events of the same category within a \textit{sequence}, and (iii)~assign the corresponding events relative to the order of occurrence within the \textit{sequence} only if the number of annotations matches.
% Non-consistent sequences
If this number does not match, we recommend to either separately consider the \textit{sequence} or to fully discard the included \textit{annotations}.
% Term SCM
In analogy to \acrshort{nnm}, we refer to this method as \acrfull{scm}.
% Expand sequence consistency according to detail in the annotations  
Please note that, relative to the degree of detail within the compared \textit{annotations}, the contained additional information (for example player identities) can be used to increase the degree of detail in \acrshort{scm}.

\section{Experiments}\label{sec:experiments}

\begin{table}[b!]
\caption{Measuring the temporal IoU of several annotators for events with a duration.}
\label{tab:tiou}
\centering
\setlength{\tabcolsep}{2pt}
\small
\fontsize{7}{10}\selectfont
\begin{tabularx}{\linewidth}{Xl|r|r}
\toprule
Annotation & Dataset &  Game Status     & Ball Possession   \\ \midrule                        
\multicolumn{1}{l|}{\multirow{2}{*}{Average Experienced}}               & \acrshort{dataset_handball}-T  &     $0.68 \pm 0.02$            &    $0.72 \pm 0.02$          \\
\multicolumn{1}{l|}{}                                           & \acrshort{dataset_soccer}-T  &   $ 0.92 \pm 0.01 $            &  $0.78 \pm 0.03$             \\
\multicolumn{1}{l|}{Inexperienced vs. Experienced}                                           &\acrshort{dataset_soccer}-T  &     $0.92 \pm 0.03$           & $0.73 \pm 0.05$  \\
\bottomrule
\end{tabularx}
\end{table}

\newcolumntype{R}[2]{%
    >{\adjustbox{angle=#1,lap=\width-(#2)}\bgroup}%
    l%
    <{\egroup}%
}
\newcommand*\rot{\multicolumn{1}{R{90}{1em}}}% no optional argument here, please!

\begin{table*}[tbh!]
\caption{Expected human and from the baseline achieved performance: Precision and Recall (in~$\%$) after applying Nearest Neighbour (NN) matching and the proposed Sequence Consistency (SC) matching for representative events at multiple hierarchy levels. Note, that an appropriate event-specific evaluation window length~$w_\text{eval}$~[s] is applied. The number of consistent events for SC matching (in~$\%$) are indicated in brackets. {\normalfont\footnotesize $^2$No \emph{ball out of field} annotations provided.}}
\label{tab:aggreement_nn_cm}
\centering
\setlength{\tabcolsep}{2.2pt}
\small
\fontsize{7}{10}\selectfont
\def\arraystretch{1.0} % 0.85
\begin{tabularx}{\textwidth}{l|l|llllllllllllllllll}
\toprule
                        & Event                    & \multicolumn{2}{l}{\emph{Status Change}}   & \multicolumn{2}{l}{\emph{Reception}}   & \multicolumn{2}{l}{\emph{Release}}   & \multicolumn{2}{l}{\emph{Int. Release}}   & \multicolumn{2}{l}{\emph{Unint. Release}}   & \multicolumn{2}{l}{\emph{Shot}}   & \multicolumn{2}{l}{\emph{Suc. Interf.}}   & \multicolumn{2}{l}{\emph{Suc. Pass}}   & \multicolumn{2}{l}{\emph{Interc. Pass}}   \\
Dataset                 & Matching                 & NN       & SC           & NN       & SC           & NN       & SC           & NN       & SC           & NN       & SC           & NN      & SC            & NN       & SC           & NN       & SC           & NN       & SC           \\ \hline 

\multirow{6}{*}{\acrshort{dataset_handball}-T} & $w_\text{eval}$            & \multicolumn{2}{c}{6.04} & \multicolumn{2}{c}{0.44} & \multicolumn{2}{c}{0.44} & \multicolumn{2}{c}{0.44} & \multicolumn{2}{c}{2.04} & \multicolumn{2}{c}{0.44} & \multicolumn{2}{c}{2.04} & \multicolumn{2}{c}{0.44} & \multicolumn{2}{c}{0.44} \\ \hline \midrule
& Num. Events               & \multicolumn{2}{c}{$135.7 \pm 23.0$}  & \multicolumn{2}{c}{$821.0 \pm 10.7$}  & \multicolumn{2}{c}{$844.7 \pm 26.8$}  & \multicolumn{2}{c}{$841.0 \pm 26.5$}  & \multicolumn{2}{c}{$3.7 \pm 0.5$}  & \multicolumn{2}{c}{$62.7 \pm 1.2$}  & \multicolumn{2}{c}{$3.0 \pm 0.8$}  & \multicolumn{2}{c}{$765.7 \pm 25.8$}  & \multicolumn{2}{c}{$6.7 \pm 0.9$}  \\ \cline{2-20}
                        & Mean Exp. (Prc.=Rec.)              
                        & 78.9         & 90.0 (40)
                        & 92.2         & 89.7 (45)
                        & 92.7         & 82.3 (33)
                        & 92.9         & 83.3 (33)
                        & 18.2         & 100 (18)
                        & 96.3         & 99.4 (93)
                        & 22.2         & 100 (22)
                        & 92.4         & 82.6 (37)    
                        & 45.0         & 100 (45)
                        \\
                        & Baseline vs. Exp. Prc.              
                        &\multicolumn{2}{c}{-}
                        & 45.6         &    0.0 (0)          
                        &  \multicolumn{2}{c}{-}           
                        &  \multicolumn{2}{c}{-}            
                        & \multicolumn{2}{c}{-}            
                        &  43.2        &   0.0 (0)            
                        &  \multicolumn{2}{c}{-}         
                        &  46.4        &    0.0 (0)          
                        &    \multicolumn{2}{c}{-} 
                        \\
                        & Baseline vs. Exp. Rec.              
                        & \multicolumn{2}{c}{-}
                        & 93.9         &    0.0 (0)          
                        &  \multicolumn{2}{c}{-}          
                        &  \multicolumn{2}{c}{-}            
                        &  \multicolumn{2}{c}{-}            
                        & 41.0         &   0.0 (0)            
                        & \multicolumn{2}{c}{-}           
                        & 91.5         &  0.0 (0)            
                        &    \multicolumn{2}{c}{-}

 \\ \hline \midrule
 
\multirow{4}{*}{\acrshort{dataset_soccer}-T}
& Num. Events               & 
                        \multicolumn{2}{c}{$113.4 \pm 3.4$ %(77)
                        } & \multicolumn{2}{c}{$362.6 \pm 5.6$ %(61)
                        } & \multicolumn{2}{c}{$550.2  \pm 13.4$ %(45)
                        }  & \multicolumn{2}{c}{$500.0  \pm 7.0$ %(49)
                        }  & \multicolumn{2}{c}{$50.2 \pm 14.2$ %(48)
                        }  & \multicolumn{2}{c}{$12.2  \pm 0.4$ %(98)
                        } & \multicolumn{2}{c}{$32.0 \pm 4.3$ %(56)
                        } & \multicolumn{2}{c}{$385.2 \pm 4.5$ %(54)
                        } & \multicolumn{2}{c}{$58.8  \pm 18.9$ %(38)$
                        }  \\ \cline{2-20}
                        % Mean Expert 
                        & Mean Exp. (Prc.=Rec.)           
                        & 95.0 & 98.7 (78)
                        & 94.9 & 94.5 (61)
                        & 95.7 & 90.8 (42)
                        & 96.2 & 93.3 (48)
                        & 62.7 & 84.4 (48)
                        & 100.0 & 100.0 (100)
                        & 68.5 & 91.8 (57)
                        & 96.0 & 94.0 (54)
                        & 60.0 & 85.0 (33)
                        \\
                        
                        % Inexperienced vs. Expert
                        & Inexp. vs Exp. Prc.     
                        & 95.9 & 98.8 (77)
                        & 90.3 & 89.7 (61)
                        & 93.4 & 90.0 (49)
                        & 94.8 & 87.9 (51)
                        & 64.1 & 78.9 (49)
                        & 92.3 & 100 (100)
                        & 64.5 & 83.3 (55)
                        & 92.6 & 93.2 (53)
                        & 65.5 & 85.5 (44)
                        \\
                        & Inexp. vs Exp. Rec.     
                        & 93.8 & 98.8 (74)
                        & 88.9 & 89.6 (60)
                        & 92.5 & 89.5 (49)
                        & 93.2 & 87.0 (50)
                        & 60.0 & 78.9 (47)
                        & 100 & 100 (92)
                        & 62.0 & 83.3 (55)
                        & 94.3 & 92.5 (54)
                        & 71.4 & 85.5 (48)
                        \\ \hline \midrule 
                        \multirow{3}{*}{\acrshort{dataset_provider}}
                         & Num. Events & \multicolumn{2}{c}{$649.0 \pm 8.0$\tablefootnote{ }} 
                         & \multicolumn{2}{c}{-}  &  \multicolumn{2}{c}{-}  & \multicolumn{2}{c}{-}  & \multicolumn{2}{c}{-}  & \multicolumn{2}{c}{$100.5 \pm 2.5$}  & \multicolumn{2}{c}{-}  & \multicolumn{2}{c}{$2428.0 \pm 24.0$}  & \multicolumn{2}{c}{-}  \\
                        \cline{2-20}
                        & Data Provider Prc.
                        &  74.7 & 71.1 (96)          
                        &  \multicolumn{2}{c}{-} 
                        &  \multicolumn{2}{c}{-} 
                        &  \multicolumn{2}{c}{-}               
                        &  \multicolumn{2}{c}{-}               
                        &  6.8 & 8.0 (73)                
                        &  \multicolumn{2}{c}{-}               
                        & 12.9         & 12.3 (65)              
                        &  \multicolumn{2}{c}{-} 
                        \\
                        & Data Provider Rec.
                        &  73.5 & 71.1 (93)           
                        &  \multicolumn{2}{c}{-} 
                        &  \multicolumn{2}{c}{-} 
                        &  \multicolumn{2}{c}{-}               
                        &  \multicolumn{2}{c}{-}              
                        &  7.1 & 8.1 (76)               
                        &  \multicolumn{2}{c}{-}               
                        & 12.6         & 12.6 (65)              
                        &  \multicolumn{2}{c}{-}  
                    
 \\ \bottomrule
\end{tabularx}
\end{table*}

We assess the quality of our proposed dataset by measuring the expected human performance~(Section~\ref{exp:aggreement}) and present a baseline classifier that only utilizes visual features~(Section~\ref{exp:baseline}). 
The %uncertainty 
quality of annotations from an official data provider is evaluated in Section~\ref{exp:data_prov_quality}.

\subsection{Assessment of Human Performance}\label{exp:aggreement}

% Terms: Agreement vs. Performance
Despite we aim to provide as clear as possible definitions for the annotated events, the complex nature of invasion games might lead to uncertain decisions during the annotation process.
According to common practice, we assess the annotation quality and, hence, expected performance of automatic solutions by measuring the average human performance on several evaluation metrics~(Section~\ref{sec:metrics}). 
In this respect, one annotator is treated as a predictor and compared to each other annotator, respectively, considered as reference. 
Consequently, the average over all reference annotators represents the individual performance of one annotator while the average across all individual performances corresponds to the average human performance. We report the average performance for experienced annotators for \acrshort{dataset_handball}-T and \acrshort{dataset_soccer}-T while we additionally assess the generality of our taxonomy by comparing the individual performance of domain experts and an inexperienced annotator for \acrshort{dataset_soccer}-T.

For events with a duration (\emph{game status}, \emph{possession}), we report the \acrshort{temporal_iou}.
To evaluate the event spotting task, a sufficient assessment of human performance requires a multitude of metrics. Similar to~\citet{sanford2020group}, we report the precision and recall by applying the \acrshort{nnm} for individual events at different levels of our proposed \textit{hierarchy}. %However, in contrast to previous works, 
We define strict but meaningful tolerance areas for each event to support the general interpretability of the results. 
% todo: Choice of eval_window -> derived from temporal agreement?
Additionally, we apply the \acrshort{scm} where we compensate for a possible varying number of sequences by adopting the sequence borders in case of a possible mismatch. 
We report precision and recall for events from consistent sequences along with the percentage of events from consistent sequences. The Appendix provides a detailed overview of each individual annotator performance.

\paragraph{Results \& Findings}

% 0. Overall performance
The overall results for events with a duration~(Table~\ref{tab:tiou}) and events with a timestamp~(Table~\ref{tab:aggreement_nn_cm}) indicate a general agreement for the discussed concepts.
% 1. Expert vs. Non-experienced results
Moreover, the minor discrepancies in the performance of the experienced and the inexperienced annotator for \acrshort{dataset_soccer}-T also indicate that a sufficient annotation of our base taxonomy does generally not require expert knowledge. This observation shows the low amount of semantic interpretation included in our proposed taxonomy. Please note that due to the asymmetry in the comparison (one inexperienced annotator as prediction and four experienced annotators as reference), for this case, the precision and recall differ in Table~\ref{tab:aggreement_nn_cm}.

% 2. tIoU + Soccer vs. Handball
%A general difference in the results for \acrshort{dataset_soccer}-T and \acrshort{dataset_handball}-T can be found in Table~\ref{tab:tiou}. 
In Table~\ref{tab:tiou}, the agreement for \textit{game status} in soccer is significantly higher than the agreement in \textit{possession}. For handball, while the results for \textit{possession} are comparable to soccer, the agreement for \textit{game status} is significantly lower. This likely originates from the rather fluent transitions between active and inactive play which complicate a clear recognition of \textit{game status change events} in handball.
% 3. PR + Soccer vs. Handball
In contrast, general similarities in the annotations for \acrshort{dataset_soccer}-T and \acrshort{dataset_handball}-T can be found in agreement for individual ball events~(Table~\ref{tab:aggreement_nn_cm}). Beneath the previously discussed differences in the ambiguity of \textit{game status}, reflected in inferior agreement of \textit{game status change events}, similar trends are observable in both sports~(limitations, i.e., for infrequent events in handball such as \textit{unintentional ball release} or \textit{successful interference}).
% 3.1. Results for individual events 
For both datasets, the \textit{hierarchical} structure positively influences the results where the highest level shows a high overall agreement which decreases when descending in the \textit{hierarchy}. This relates to the similarly increasing level of included semantic information~(see Section~\ref{subsec:characteristics}) complicating the annotation. However, this general observation does not translate to each particular event in the taxonomy. 
%For instance, \textit{intentional ball releases} such as a \textit{successful pass} show a high agreement even at low \textit{hierarchy} levels.
%This complies with the intuition that it is easier to recognize a present intention~(or success) than to clearly decide on a (possibly) missing one.% 3.1.2. Results for \acrshort{nnd \acrshort{scm}

The results for \acrshort{scm} provide a valuable extension to the informative value of \acrshort{nnm}, i.e., to detect the positive bias~(Section~\ref{sec:metrics}). For instance, the \textit{successful pass} for \acrshort{dataset_handball}-T shows a general high agreement. However, a positive bias in this metric can be recognized regarding the comparatively low amount of sequence-consistent events~(in brackets). These differences are probably caused by the high frequency of \textit{successful passes} in handball and the connected issues with assignment, detailed in Section~\ref{sec:metrics}.  

% 4. Example for an uncertain situation (uncertain ball control/clearance)
% 4.1. Semantic Misclassification
% a) Heading duel 
Typical misclassifications are often related to the assignment of intention. For ambiguous situations~(see Figure~\ref{fig:example_annotations}), this assignment can be difficult and largely depending on the outcome. For instance, if a played ball lands close to a teammate the situation will rather be annotated as \textit{intentional ball release}. However, this does not comply with the concept of intention that needs to be distinguished in the moment of the execution. Yet, due to the complex nature of invasion games, even the player who played the ball might not give a definite answer.          
% 4.2. Temporal Misclassification
% a) Due to Cuts or Replays
A different type of error are temporal mismatches~(such as delays). While generally excluded from the annotation, still, a common source for these temporal differences are cuts, replays, or close-ups in the video data. As we aim to include the majority of events
%albeit occurring the replays 
if the action on the pitch can be derived from the general situation~(i.e., a replay only overlaps with a small fraction of an event), a common source of error are different event times. This is especially relevant for \textit{game status change events} where cuts and replays commonly occur.

\subsection{Vision-based Baseline}\label{exp:baseline}
To present some potential outputs of an automated classifier model, we create a baseline that only uses visual features from the raw input video to spot events.
Due to the lack of existing end-to-end solutions for event spotting and density of events (approx. each second in \acrshort{dataset_handball}), we follow common practice, where first a classifier is trained on short clips and then a sliding window is applied to produce frame-wise output~(e.g., feature vectors or class probabilities). 
We follow \cite{sanford2020group} and directly apply \acrshort{nms} to the predicted class probabilities to suppress several positive predictions around the same event.

\subsubsection{Setup for Video Chunk Classification}
For the model, we choose an Inflated 3D ConvNet~\emph{I3D}~\cite{carreira2017quo, NonLocal2018} with a \emph{ResNet-50} as backbone which is pre-trained on \emph{Kinetics400}~\cite{kay2017kinetics}. 
We select three classes~(\emph{reception}, \emph{successful pass}, and \emph{shot})~(plus a background event). We train one model for \acrshort{dataset_handball} on the entire~(spatial) visual content with fixed input dimension of $Tx3x256x456$.
Short clips~($T=32$ frames), centered around the annotation, are created to cover temporal context. For the background event, all remaining clips are taken with a stride of 15 frames. 
Temporal resolution is halved to during training~(i.e., 15\,fps for \acrshort{dataset_handball}).
For remaining details we refer to the Appendix.
The model with the lowest validation loss is selected for the event spotting task.

\subsubsection{Evaluating the Event Spotting Task}
We collect all predicted probabilities at each frame using a sliding window and apply \acrshort{nms} on validated event-specific filter lengths~$w^e_\text{nms}$. 
As several events can occur at the same time, for each event~$e$ a confidence threshold~$\tau_e$ is estimated.
Both hyper-parameters are optimized for each event on the $F_1$ score with \acrshort{nnm} using a grid search on the training dataset. We use the same search space as \citet{sanford2020group}.

Results are reported in Table~\ref{tab:aggreement_nn_cm} where precision and recall are calculated considering the expert annotation with the highest human performance as ground-truth.
% two or three sentences w.r.t. results
Despite the limited amount of training data, the baseline demonstrates that our proposed datasets are suitable for benchmarking on-ball events.
We qualitatively observe that an excessive number of positive predictions in spite of \acrshort{nms} causes bad performance using \acrshort{scm}, which is only partly visible when using \acrshort{nnm}. 
This confirms the need for the proposed metric and identifies the error cases of the baseline.
The model achieves sufficiently robust recognition performance with temporal centered ground-truth events, it predicts the actual event with high confidence when an ground-truth event in the sliding window is not centered.
We refer to future work~(1)~to improve the visual model for instance with hard-negative sample mining, or temporal pooling~\cite{giancola2021temporally} and~(2)~for the usage of multimodal data~(e.g., \cite{vanderplaetse2020improved}).

\subsection{Annotation Quality of Data Providers}\label{exp:data_prov_quality}

% Semantic and temporal errors of existing data
As previously discussed, annotations (in soccer) are frequently obtained from data providers that are not bound to fulfill any requirements or to meet a common gold standard.
% Opta data
To this end, we explore the quality of a data provider on the exemplary \acrfull{dataset_provider} which contains four matches of a first European soccer league from the 2014/2015 season. Here, we avoid an examination of semantically complex events like \textit{successful interference} % where differences in the definition may constrain the obtained results.
%Specifically, we 
and perform an examination of the \textit{successful pass}, \textit{shot}, and \textit{game status changing events} where we find the largest compliance with the data-provider event catalog ~\cite{liu2013reliability}. To obtain a reference, we instruct a domain expert to acquire frame-wise accurate annotations by watching unedited recordings of the matches. 
% Sequence consistency
Similar to the previous experiments, we compute precision and recall while we account for differences in the number of total annotated events by application of \acrshort{scm}~(with specific consideration of passing player identities for passes). The results are given in Table~\ref{tab:aggreement_nn_cm} and a representative example is displayed in Figure~\ref{fig:example_annotations}.
% Observation
We observe a low agreement between the precise expert and the data-provider annotation~(compared to results for~\acrshort{dataset_soccer}). While due to the consideration of player identities, slightly more \textit{successful pass} events are consistent, the agreement for \acrshort{scm} is also poor.
% Explanation
This is caused by a general imprecision in the data-provider annotation. This imprecision likely originates from the real-time manual annotation which data providers demand. The human annotators are instructed to collect temporal (and spatial) characteristics of specific events while simultaneously deciding on a corresponding event type from a rather large range of \textit{high-level} event catalog ~\cite{liu2013reliability}. 
These results reveal the need for exact definitions and annotation guidelines and emphasize the value of automatic solutions.
We intend to show with this exploratory experiment that, the quality of the annotations provided should be taken into account depending on the targeted application. 
Of course, we cannot draw conclusions about quality of other data and seasons based on this case study.

\section{Conclusions}\label{sec:conclusion}
% Problems
In this paper, we have addressed the real-world task of fine-grained event detection and spotting in invasion games. 
While prior work already presented automatic methods for the detection of individual sport-specific events with focus on soccer, they lacked objective event definitions and complete description for invasion games.
Despite the wide range of examined events, their complexity, and ambiguity, the data quality had not been investigated making the assessment of automatic approaches difficult. Even current evaluation metrics are inconsistent. 
% Contribution 
Therefore, we have contributed a \textit{hierarchical} taxonomy that enables a \textit{minimal} and \textit{objective} annotation and is \textit{modular expendable} to fit the needs of various invasion games. 
In addition, we released two multimodal datasets with gold standard event annotations~(soccer and handball).
% Evaluation -> High agreements w/o/ expert knowledge
Extensive evaluation have validated the taxonomy as well as the quality of our two benchmark datasets while a comparison with data-provider annotations revealed advantages in annotation quality.
The results have shown that high agreement can be achieved even without domain knowledge. In addition, the \textit{hierarchical} approach demonstrates that (semantically) complex events can be propagated to a shared parent event to reach an increase in agreement.

With the presented taxonomy, datasets, and baseline, we create a foundation for the design and the benchmarking of upcoming automatic approaches for the spotting of on-ball events.
Also, other domains that work with video, positional, and event data, could benefit from the taxonomy and the datasets introduced in this paper.
In the future, we plan to integrate non-on-ball events into the taxonomy and to exploit \textit{hierarchical} information and attention to the ball position during training of a deep model.

\section*{Acknowledgement}
% This project (project number: <project_number>) has received funding from the German Federal Ministry of Education and Research (BMBF - Bundesministerium für Bildung und Forschung) under <Förderkennzeichen>.
This project has received funding from the German Federal Ministry of Education and Research (BMBF -- Bundesministerium für Bildung und Forschung) under 01IS20021A, 01IS20021B, and 01IS20021C.
This research was supported by a grant from the German Research Council (DFG, Deutsche Forschungsgemeinschaft) to DM (grant ME~2678/30.1).

\bibliographystyle{ACM-Reference-Format}
\bibliography{references}


\begin{thebibliography}{60}

%%% the Web address.

\ifx \showCODEN    \undefined \def \showCODEN     #1{\unskip}     \fi
\ifx \showDOI      \undefined \def \showDOI       #1{#1}\fi
\ifx \showISBNx    \undefined \def \showISBNx     #1{\unskip}     \fi
\ifx \showISBNxiii \undefined \def \showISBNxiii  #1{\unskip}     \fi
\ifx \showISSN     \undefined \def \showISSN      #1{\unskip}     \fi
\ifx \showLCCN     \undefined \def \showLCCN      #1{\unskip}     \fi
\ifx \shownote     \undefined \def \shownote      #1{#1}          \fi
\ifx \showarticletitle \undefined \def \showarticletitle #1{#1}   \fi
\ifx \showURL      \undefined \def \showURL       {\relax}        \fi
% invisible to TeX
\providecommand\bibfield[2]{#2}
\providecommand\bibinfo[2]{#2}
\providecommand\natexlab[1]{#1}
\providecommand\showeprint[2][]{arXiv:#2}

\bibitem[\protect\citeauthoryear{??}{IFA}{2021}]%
        {IFAB}
 \bibinfo{year}{2021}\natexlab{}.
\newblock \showarticletitle{{IFAB Laws of soccer}}.
\newblock  (\bibinfo{year}{2021}).
\newblock
\urldef\tempurl%
\url{https://www.theifab.com/laws}
\showURL{%
\tempurl}


\bibitem[\protect\citeauthoryear{??}{IHF}{2021}]%
        {IHF}
 \bibinfo{year}{2021}\natexlab{}.
\newblock \showarticletitle{{IHF Laws of handball}}.
\newblock  (\bibinfo{year}{2021}).
\newblock
\urldef\tempurl%
\url{https://www.ihf.info/regulations-documents/361?selected=Rules\%20of\%20the\%20Game}
\showURL{%
\tempurl}


\bibitem[\protect\citeauthoryear{??}{opt}{2021}]%
        {opta}
 \bibinfo{year}{2021}\natexlab{}.
\newblock \bibinfo{title}{OptaSports}.
\newblock \bibinfo{howpublished}{\url{https://www.optasports.com}}.
\newblock
\newblock
\shownote{Accessed: 2021-01-19.}


\bibitem[\protect\citeauthoryear{??}{Spo}{2021}]%
        {SportsML}
 \bibinfo{year}{2021}\natexlab{}.
\newblock \showarticletitle{{SportsML open standard for sports data}}.
\newblock  (\bibinfo{year}{2021}).
\newblock
\urldef\tempurl%
\url{https://iptc.org/standards/sportsml-g2/}
\showURL{%
\tempurl}


\bibitem[\protect\citeauthoryear{??}{sta}{2021}]%
        {stats}
 \bibinfo{year}{2021}\natexlab{}.
\newblock \bibinfo{title}{STATS}.
\newblock \bibinfo{howpublished}{\url{https://www.statsperform.com/}}.
\newblock
\newblock
\shownote{Accessed: 2021-01-19.}


\bibitem[\protect\citeauthoryear{??}{wys}{2021}]%
        {wyscout}
 \bibinfo{year}{2021}\natexlab{}.
\newblock \bibinfo{title}{wyscout}.
\newblock \bibinfo{howpublished}{\url{https://wyscout.com/}}.
\newblock
\newblock
\shownote{Accessed: 2021-01-19.}


\bibitem[\protect\citeauthoryear{Burger, Rogulj, Foreti{\'c}, and
  {\v{C}}avala}{Burger et~al\mbox{.}}{2013}]%
        {burger2013analysis}
\bibfield{author}{\bibinfo{person}{Ante Burger}, \bibinfo{person}{Nenad
  Rogulj}, \bibinfo{person}{Nikola Foreti{\'c}}, {and}
  \bibinfo{person}{Marijana {\v{C}}avala}.} \bibinfo{year}{2013}\natexlab{}.
\newblock \showarticletitle{Analysis of rebounded balls in a team handball
  match}.
\newblock \bibinfo{journal}{\emph{SportLogia}} \bibinfo{volume}{9},
  \bibinfo{number}{1} (\bibinfo{year}{2013}), \bibinfo{pages}{53--58}.
\newblock


\bibitem[\protect\citeauthoryear{Caba~Heilbron, Escorcia, Ghanem, and
  Carlos~Niebles}{Caba~Heilbron et~al\mbox{.}}{2015}]%
        {caba2015activitynet}
\bibfield{author}{\bibinfo{person}{Fabian Caba~Heilbron},
  \bibinfo{person}{Victor Escorcia}, \bibinfo{person}{Bernard Ghanem}, {and}
  \bibinfo{person}{Juan Carlos~Niebles}.} \bibinfo{year}{2015}\natexlab{}.
\newblock \showarticletitle{Activitynet: A large-scale video benchmark for
  human activity understanding}. In \bibinfo{booktitle}{\emph{Proceedings of
  the ieee conference on computer vision and pattern recognition}}.
  \bibinfo{pages}{961--970}.
\newblock


\bibitem[\protect\citeauthoryear{Camerino, Chaverri, Anguera, and
  Jonsson}{Camerino et~al\mbox{.}}{2012}]%
        {camerino2012dynamics}
\bibfield{author}{\bibinfo{person}{Oleguer~Foguet Camerino},
  \bibinfo{person}{Javier Chaverri}, \bibinfo{person}{M~Teresa Anguera}, {and}
  \bibinfo{person}{Gudberg~K Jonsson}.} \bibinfo{year}{2012}\natexlab{}.
\newblock \showarticletitle{Dynamics of the game in soccer: Detection of
  T-patterns}.
\newblock \bibinfo{journal}{\emph{European Journal of Sport Science}}
  \bibinfo{volume}{12}, \bibinfo{number}{3} (\bibinfo{year}{2012}),
  \bibinfo{pages}{216--224}.
\newblock


\bibitem[\protect\citeauthoryear{Carreira and Zisserman}{Carreira and
  Zisserman}{2017}]%
        {carreira2017quo}
\bibfield{author}{\bibinfo{person}{Joao Carreira} {and} \bibinfo{person}{Andrew
  Zisserman}.} \bibinfo{year}{2017}\natexlab{}.
\newblock \showarticletitle{Quo vadis, action recognition? a new model and the
  kinetics dataset}. In \bibinfo{booktitle}{\emph{proceedings of the IEEE
  Conference on Computer Vision and Pattern Recognition}}.
  \bibinfo{pages}{6299--6308}.
\newblock


\bibitem[\protect\citeauthoryear{Casal, Maneiro, Ard{\'a}, Mar{\'\i}, and
  Losada}{Casal et~al\mbox{.}}{2017}]%
        {casal2017possession}
\bibfield{author}{\bibinfo{person}{Claudio~A Casal}, \bibinfo{person}{Rub{\'e}n
  Maneiro}, \bibinfo{person}{Toni Ard{\'a}}, \bibinfo{person}{Francisco~J
  Mar{\'\i}}, {and} \bibinfo{person}{Jos{\'e}~L Losada}.}
  \bibinfo{year}{2017}\natexlab{}.
\newblock \showarticletitle{Possession zone as a performance indicator in
  football. The game of the best teams}.
\newblock \bibinfo{journal}{\emph{Frontiers in psychology}}
  \bibinfo{volume}{8} (\bibinfo{year}{2017}), \bibinfo{pages}{1176}.
\newblock


\bibitem[\protect\citeauthoryear{Chacoma, Almeira, Perotti, and
  Billoni}{Chacoma et~al\mbox{.}}{2020}]%
        {chacoma2020modeling}
\bibfield{author}{\bibinfo{person}{A Chacoma}, \bibinfo{person}{N Almeira},
  \bibinfo{person}{JI Perotti}, {and} \bibinfo{person}{OV Billoni}.}
  \bibinfo{year}{2020}\natexlab{}.
\newblock \showarticletitle{Modeling ball possession dynamics in the game of
  football}.
\newblock \bibinfo{journal}{\emph{Physical Review E}} \bibinfo{volume}{102},
  \bibinfo{number}{4} (\bibinfo{year}{2020}), \bibinfo{pages}{042120}.
\newblock


\bibitem[\protect\citeauthoryear{Cioppa, Deliege, Giancola, Ghanem,
  Droogenbroeck, Gade, and Moeslund}{Cioppa et~al\mbox{.}}{2020}]%
        {cioppa2020context}
\bibfield{author}{\bibinfo{person}{Anthony Cioppa}, \bibinfo{person}{Adrien
  Deliege}, \bibinfo{person}{Silvio Giancola}, \bibinfo{person}{Bernard
  Ghanem}, \bibinfo{person}{Marc~Van Droogenbroeck}, \bibinfo{person}{Rikke
  Gade}, {and} \bibinfo{person}{Thomas~B Moeslund}.}
  \bibinfo{year}{2020}\natexlab{}.
\newblock \showarticletitle{A context-aware loss function for action spotting
  in soccer videos}. In \bibinfo{booktitle}{\emph{Proceedings of the IEEE/CVF
  Conference on Computer Vision and Pattern Recognition}}.
  \bibinfo{pages}{13126--13136}.
\newblock


\bibitem[\protect\citeauthoryear{Decroos, Bransen, Van~Haaren, and
  Davis}{Decroos et~al\mbox{.}}{2019}]%
        {decroos2019actions}
\bibfield{author}{\bibinfo{person}{Tom Decroos}, \bibinfo{person}{Lotte
  Bransen}, \bibinfo{person}{Jan Van~Haaren}, {and} \bibinfo{person}{Jesse
  Davis}.} \bibinfo{year}{2019}\natexlab{}.
\newblock \showarticletitle{Actions speak louder than goals: Valuing player
  actions in soccer}. In \bibinfo{booktitle}{\emph{Proceedings of the 25th ACM
  SIGKDD International Conference on Knowledge Discovery \& Data Mining}}.
  \bibinfo{pages}{1851--1861}.
\newblock


\bibitem[\protect\citeauthoryear{Deli{\`e}ge, Cioppa, Giancola, Seikavandi,
  Dueholm, Nasrollahi, Ghanem, Moeslund, and Van~Droogenbroeck}{Deli{\`e}ge
  et~al\mbox{.}}{2020}]%
        {deliege2020soccernet}
\bibfield{author}{\bibinfo{person}{Adrien Deli{\`e}ge},
  \bibinfo{person}{Anthony Cioppa}, \bibinfo{person}{Silvio Giancola},
  \bibinfo{person}{Meisam~J Seikavandi}, \bibinfo{person}{Jacob~V Dueholm},
  \bibinfo{person}{Kamal Nasrollahi}, \bibinfo{person}{Bernard Ghanem},
  \bibinfo{person}{Thomas~B Moeslund}, {and} \bibinfo{person}{Marc
  Van~Droogenbroeck}.} \bibinfo{year}{2020}\natexlab{}.
\newblock \showarticletitle{SoccerNet-v2: A Dataset and Benchmarks for Holistic
  Understanding of Broadcast Soccer Videos}.
\newblock \bibinfo{journal}{\emph{arXiv preprint arXiv:2011.13367}}
  (\bibinfo{year}{2020}).
\newblock


\bibitem[\protect\citeauthoryear{Dodge, Weibel, and Lautensch{\"u}tz}{Dodge
  et~al\mbox{.}}{2008}]%
        {dodge2008towards}
\bibfield{author}{\bibinfo{person}{Somayeh Dodge}, \bibinfo{person}{Robert
  Weibel}, {and} \bibinfo{person}{Anna-Katharina Lautensch{\"u}tz}.}
  \bibinfo{year}{2008}\natexlab{}.
\newblock \showarticletitle{Towards a taxonomy of movement patterns}.
\newblock \bibinfo{journal}{\emph{Information visualization}}
  \bibinfo{volume}{7}, \bibinfo{number}{3-4} (\bibinfo{year}{2008}),
  \bibinfo{pages}{240--252}.
\newblock


\bibitem[\protect\citeauthoryear{Feichtenhofer, Fan, Malik, and
  He}{Feichtenhofer et~al\mbox{.}}{2019}]%
        {feichtenhofer2019slowfast}
\bibfield{author}{\bibinfo{person}{Christoph Feichtenhofer},
  \bibinfo{person}{Haoqi Fan}, \bibinfo{person}{Jitendra Malik}, {and}
  \bibinfo{person}{Kaiming He}.} \bibinfo{year}{2019}\natexlab{}.
\newblock \showarticletitle{Slowfast networks for video recognition}. In
  \bibinfo{booktitle}{\emph{Proceedings of the IEEE/CVF International
  Conference on Computer Vision}}. \bibinfo{pages}{6202--6211}.
\newblock


\bibitem[\protect\citeauthoryear{Feng, Song, Yu, Chen, Zhao, He, and Guan}{Feng
  et~al\mbox{.}}{2020}]%
        {feng2020sset}
\bibfield{author}{\bibinfo{person}{Na Feng}, \bibinfo{person}{Zikai Song},
  \bibinfo{person}{Junqing Yu}, \bibinfo{person}{Yi-Ping~Phoebe Chen},
  \bibinfo{person}{Yizhu Zhao}, \bibinfo{person}{Yunfeng He}, {and}
  \bibinfo{person}{Tao Guan}.} \bibinfo{year}{2020}\natexlab{}.
\newblock \showarticletitle{SSET: a dataset for shot segmentation, event
  detection, player tracking in soccer videos}.
\newblock \bibinfo{journal}{\emph{Multimedia Tools and Applications}}
  \bibinfo{volume}{79}, \bibinfo{number}{39} (\bibinfo{year}{2020}),
  \bibinfo{pages}{28971--28992}.
\newblock


\bibitem[\protect\citeauthoryear{Fernandes, Camerino, Garganta, Pereira, and
  Barreira}{Fernandes et~al\mbox{.}}{2019}]%
        {fernandes2019design}
\bibfield{author}{\bibinfo{person}{Tiago Fernandes}, \bibinfo{person}{Oleguer
  Camerino}, \bibinfo{person}{Julio Garganta}, \bibinfo{person}{Rogerio
  Pereira}, {and} \bibinfo{person}{Daniel Barreira}.}
  \bibinfo{year}{2019}\natexlab{}.
\newblock \showarticletitle{Design and validation of an observational
  instrument for defence in soccer based on the dynamical systems theory}.
\newblock \bibinfo{journal}{\emph{International Journal of Sports Science \&
  Coaching}} \bibinfo{volume}{14}, \bibinfo{number}{2} (\bibinfo{year}{2019}),
  \bibinfo{pages}{138--152}.
\newblock


\bibitem[\protect\citeauthoryear{Fern{\'a}ndez and Bornn}{Fern{\'a}ndez and
  Bornn}{2020}]%
        {fernandez2020soccermap}
\bibfield{author}{\bibinfo{person}{Javier Fern{\'a}ndez} {and}
  \bibinfo{person}{Luke Bornn}.} \bibinfo{year}{2020}\natexlab{}.
\newblock \showarticletitle{SoccerMap: A Deep Learning Architecture for
  Visually-Interpretable Analysis in Soccer}.
\newblock \bibinfo{journal}{\emph{arXiv preprint arXiv:2010.10202}}
  (\bibinfo{year}{2020}).
\newblock


\bibitem[\protect\citeauthoryear{Giancola, Amine, Dghaily, and Ghanem}{Giancola
  et~al\mbox{.}}{2018}]%
        {giancola2018soccernet}
\bibfield{author}{\bibinfo{person}{Silvio Giancola},
  \bibinfo{person}{Mohieddine Amine}, \bibinfo{person}{Tarek Dghaily}, {and}
  \bibinfo{person}{Bernard Ghanem}.} \bibinfo{year}{2018}\natexlab{}.
\newblock \showarticletitle{Soccernet: A scalable dataset for action spotting
  in soccer videos}. In \bibinfo{booktitle}{\emph{Proceedings of the IEEE
  Conference on Computer Vision and Pattern Recognition Workshops}}.
  \bibinfo{pages}{1711--1721}.
\newblock


\bibitem[\protect\citeauthoryear{Giancola and Ghanem}{Giancola and
  Ghanem}{2021}]%
        {giancola2021temporally}
\bibfield{author}{\bibinfo{person}{Silvio Giancola} {and}
  \bibinfo{person}{Bernard Ghanem}.} \bibinfo{year}{2021}\natexlab{}.
\newblock \showarticletitle{Temporally-Aware Feature Pooling for Action
  Spotting in Soccer Broadcasts}.
\newblock \bibinfo{journal}{\emph{arXiv preprint arXiv:2104.06779}}
  (\bibinfo{year}{2021}).
\newblock


\bibitem[\protect\citeauthoryear{Hu, Feng, and Liu}{Hu et~al\mbox{.}}{2020}]%
        {hu2020hfnet}
\bibfield{author}{\bibinfo{person}{Lianyu Hu}, \bibinfo{person}{Lin Feng},
  {and} \bibinfo{person}{Shenglan Liu}.} \bibinfo{year}{2020}\natexlab{}.
\newblock \showarticletitle{HFNet: A Novel Model for Human Focused Sports
  Action Recognition}. In \bibinfo{booktitle}{\emph{Proceedings of the 3rd
  International Workshop on Multimedia Content Analysis in Sports}}.
  \bibinfo{pages}{35--43}.
\newblock


\bibitem[\protect\citeauthoryear{Hughes and Bartlett}{Hughes and
  Bartlett}{2002}]%
        {Hughes2002}
\bibfield{author}{\bibinfo{person}{Mike~D. Hughes} {and}
  \bibinfo{person}{Roger~M. Bartlett}.} \bibinfo{year}{2002}\natexlab{}.
\newblock \showarticletitle{The Use of Performance Indicators in Performance
  Analysis}.
\newblock \bibinfo{journal}{\emph{Journal of Sports Sciences}}
  \bibinfo{volume}{20}, \bibinfo{number}{10} (\bibinfo{date}{Jan.}
  \bibinfo{year}{2002}), \bibinfo{pages}{739--754}.
\newblock
\showISSN{0264-0414, 1466-447X}
\urldef\tempurl%
\url{https://doi.org/10.1080/026404102320675602}
\showDOI{\tempurl}


\bibitem[\protect\citeauthoryear{Ibrahim, Muralidharan, Deng, Vahdat, and
  Mori}{Ibrahim et~al\mbox{.}}{2016}]%
        {ibrahim2016hierarchical}
\bibfield{author}{\bibinfo{person}{Mostafa~S Ibrahim},
  \bibinfo{person}{Srikanth Muralidharan}, \bibinfo{person}{Zhiwei Deng},
  \bibinfo{person}{Arash Vahdat}, {and} \bibinfo{person}{Greg Mori}.}
  \bibinfo{year}{2016}\natexlab{}.
\newblock \showarticletitle{A hierarchical deep temporal model for group
  activity recognition}. In \bibinfo{booktitle}{\emph{Proceedings of the IEEE
  Conference on Computer Vision and Pattern Recognition}}.
  \bibinfo{pages}{1971--1980}.
\newblock


\bibitem[\protect\citeauthoryear{Jiang, Lu, and Xue}{Jiang
  et~al\mbox{.}}{2016}]%
        {jiang2016automatic}
\bibfield{author}{\bibinfo{person}{Haohao Jiang}, \bibinfo{person}{Yao Lu},
  {and} \bibinfo{person}{Jing Xue}.} \bibinfo{year}{2016}\natexlab{}.
\newblock \showarticletitle{Automatic soccer video event detection based on a
  deep neural network combined cnn and rnn}. In \bibinfo{booktitle}{\emph{2016
  IEEE 28th International Conference on Tools with Artificial Intelligence
  (ICTAI)}}. IEEE, \bibinfo{pages}{490--494}.
\newblock


\bibitem[\protect\citeauthoryear{Jiang, Cui, Chen, Wang, and Xu}{Jiang
  et~al\mbox{.}}{2020}]%
        {jiang2020soccerdb}
\bibfield{author}{\bibinfo{person}{Yudong Jiang}, \bibinfo{person}{Kaixu Cui},
  \bibinfo{person}{Leilei Chen}, \bibinfo{person}{Canjin Wang}, {and}
  \bibinfo{person}{Changliang Xu}.} \bibinfo{year}{2020}\natexlab{}.
\newblock \showarticletitle{Soccerdb: A large-scale database for comprehensive
  video understanding}. In \bibinfo{booktitle}{\emph{Proceedings of the 3rd
  International Workshop on Multimedia Content Analysis in Sports}}.
  \bibinfo{pages}{1--8}.
\newblock


\bibitem[\protect\citeauthoryear{Jones, James, and Mellalieu}{Jones
  et~al\mbox{.}}{2004}]%
        {jones2004possession}
\bibfield{author}{\bibinfo{person}{PD Jones}, \bibinfo{person}{Nic James},
  {and} \bibinfo{person}{Stephen~D Mellalieu}.}
  \bibinfo{year}{2004}\natexlab{}.
\newblock \showarticletitle{Possession as a performance indicator in soccer.}
\newblock \bibinfo{journal}{\emph{International Journal of Performance Analysis
  in Sport}} \bibinfo{volume}{4}, \bibinfo{number}{1} (\bibinfo{year}{2004}),
  \bibinfo{pages}{98--102}.
\newblock


\bibitem[\protect\citeauthoryear{Karimi, Toosi, and Akhaee}{Karimi
  et~al\mbox{.}}{2021}]%
        {karimi2021soccer}
\bibfield{author}{\bibinfo{person}{Ali Karimi}, \bibinfo{person}{Ramin Toosi},
  {and} \bibinfo{person}{Mohammad~Ali Akhaee}.}
  \bibinfo{year}{2021}\natexlab{}.
\newblock \showarticletitle{Soccer Event Detection Using Deep Learning}.
\newblock \bibinfo{journal}{\emph{arXiv preprint arXiv:2102.04331}}
  (\bibinfo{year}{2021}).
\newblock


\bibitem[\protect\citeauthoryear{Kay, Carreira, Simonyan, Zhang, Hillier,
  Vijayanarasimhan, Viola, Green, Back, Natsev, et~al\mbox{.}}{Kay
  et~al\mbox{.}}{2017}]%
        {kay2017kinetics}
\bibfield{author}{\bibinfo{person}{Will Kay}, \bibinfo{person}{Joao Carreira},
  \bibinfo{person}{Karen Simonyan}, \bibinfo{person}{Brian Zhang},
  \bibinfo{person}{Chloe Hillier}, \bibinfo{person}{Sudheendra
  Vijayanarasimhan}, \bibinfo{person}{Fabio Viola}, \bibinfo{person}{Tim
  Green}, \bibinfo{person}{Trevor Back}, \bibinfo{person}{Paul Natsev},
  {et~al\mbox{.}}} \bibinfo{year}{2017}\natexlab{}.
\newblock \showarticletitle{The kinetics human action video dataset}.
\newblock \bibinfo{journal}{\emph{arXiv preprint arXiv:1705.06950}}
  (\bibinfo{year}{2017}).
\newblock


\bibitem[\protect\citeauthoryear{Khaustov and Mozgovoy}{Khaustov and
  Mozgovoy}{2020}]%
        {khaustov2020recognizing}
\bibfield{author}{\bibinfo{person}{Victor Khaustov} {and}
  \bibinfo{person}{Maxim Mozgovoy}.} \bibinfo{year}{2020}\natexlab{}.
\newblock \showarticletitle{Recognizing Events in Spatiotemporal Soccer Data}.
\newblock \bibinfo{journal}{\emph{Applied Sciences}} \bibinfo{volume}{10},
  \bibinfo{number}{22} (\bibinfo{year}{2020}), \bibinfo{pages}{8046}.
\newblock


\bibitem[\protect\citeauthoryear{Kim, James, Parmar, Ali, and
  Vu{\v{c}}kovi{\'c}}{Kim et~al\mbox{.}}{2019}]%
        {kim2019attacking}
\bibfield{author}{\bibinfo{person}{Jongwon Kim}, \bibinfo{person}{Nic James},
  \bibinfo{person}{Nimai Parmar}, \bibinfo{person}{Besim Ali}, {and}
  \bibinfo{person}{Goran Vu{\v{c}}kovi{\'c}}.} \bibinfo{year}{2019}\natexlab{}.
\newblock \showarticletitle{The attacking process in football: a taxonomy for
  classifying how teams create goal scoring opportunities using a case study of
  crystal Palace FC}.
\newblock \bibinfo{journal}{\emph{Frontiers in psychology}}
  \bibinfo{volume}{10} (\bibinfo{year}{2019}), \bibinfo{pages}{2202}.
\newblock


\bibitem[\protect\citeauthoryear{Lago-Pe{\~n}as, Lago-Ballesteros, Dellal, and
  G{\'o}mez}{Lago-Pe{\~n}as et~al\mbox{.}}{2010}]%
        {lago2010game}
\bibfield{author}{\bibinfo{person}{Carlos Lago-Pe{\~n}as},
  \bibinfo{person}{Joaqu{\'\i}n Lago-Ballesteros}, \bibinfo{person}{Alexandre
  Dellal}, {and} \bibinfo{person}{Maite G{\'o}mez}.}
  \bibinfo{year}{2010}\natexlab{}.
\newblock \showarticletitle{Game-related statistics that discriminated winning,
  drawing and losing teams from the Spanish soccer league}.
\newblock \bibinfo{journal}{\emph{Journal of sports science \& medicine}}
  \bibinfo{volume}{9}, \bibinfo{number}{2} (\bibinfo{year}{2010}),
  \bibinfo{pages}{288}.
\newblock


\bibitem[\protect\citeauthoryear{Lamas, Barrera, Otranto, and
  Ugrinowitsch}{Lamas et~al\mbox{.}}{2014}]%
        {lamas2014invasion}
\bibfield{author}{\bibinfo{person}{Leonardo Lamas}, \bibinfo{person}{Junior
  Barrera}, \bibinfo{person}{Guilherme Otranto}, {and} \bibinfo{person}{Carlos
  Ugrinowitsch}.} \bibinfo{year}{2014}\natexlab{}.
\newblock \showarticletitle{Invasion team sports: strategy and match modeling}.
\newblock \bibinfo{journal}{\emph{International Journal of Performance Analysis
  in Sport}} \bibinfo{volume}{14}, \bibinfo{number}{1} (\bibinfo{year}{2014}),
  \bibinfo{pages}{307--329}.
\newblock


\bibitem[\protect\citeauthoryear{Lin, Liu, Li, Ding, and Wen}{Lin
  et~al\mbox{.}}{2019}]%
        {lin2019bmn}
\bibfield{author}{\bibinfo{person}{Tianwei Lin}, \bibinfo{person}{Xiao Liu},
  \bibinfo{person}{Xin Li}, \bibinfo{person}{Errui Ding}, {and}
  \bibinfo{person}{Shilei Wen}.} \bibinfo{year}{2019}\natexlab{}.
\newblock \showarticletitle{Bmn: Boundary-matching network for temporal action
  proposal generation}. In \bibinfo{booktitle}{\emph{Proceedings of the
  IEEE/CVF International Conference on Computer Vision}}.
  \bibinfo{pages}{3889--3898}.
\newblock


\bibitem[\protect\citeauthoryear{Link and Hoernig}{Link and Hoernig}{2017}]%
        {link2017individual}
\bibfield{author}{\bibinfo{person}{Daniel Link} {and} \bibinfo{person}{Martin
  Hoernig}.} \bibinfo{year}{2017}\natexlab{}.
\newblock \showarticletitle{Individual ball possession in soccer}.
\newblock \bibinfo{journal}{\emph{PloS one}} \bibinfo{volume}{12},
  \bibinfo{number}{7} (\bibinfo{year}{2017}), \bibinfo{pages}{e0179953}.
\newblock


\bibitem[\protect\citeauthoryear{Liu, Hopkins, Gómez, and Molinuevo}{Liu
  et~al\mbox{.}}{2013}]%
        {liu2013reliability}
\bibfield{author}{\bibinfo{person}{Hongyou Liu}, \bibinfo{person}{Will
  Hopkins}, \bibinfo{person}{A~Miguel Gómez}, {and} \bibinfo{person}{S~Javier
  Molinuevo}.} \bibinfo{year}{2013}\natexlab{}.
\newblock \showarticletitle{Inter-operator reliability of live football match
  statistics from {OPTA} {Sportsdata}}.
\newblock \bibinfo{journal}{\emph{Int J Perform Anal Sport}}
  \bibinfo{volume}{13}, \bibinfo{number}{3} (\bibinfo{year}{2013}),
  \bibinfo{pages}{803--821}.
\newblock
\newblock
\shownote{Publisher: Taylor \& Francis.}


\bibitem[\protect\citeauthoryear{Liu, Lu, Lei, Zhang, Wang, Huang, and
  Wang}{Liu et~al\mbox{.}}{2017}]%
        {liu2017soccer}
\bibfield{author}{\bibinfo{person}{Tingxi Liu}, \bibinfo{person}{Yao Lu},
  \bibinfo{person}{Xiaoyu Lei}, \bibinfo{person}{Lijing Zhang},
  \bibinfo{person}{Haoyu Wang}, \bibinfo{person}{Wei Huang}, {and}
  \bibinfo{person}{Zijian Wang}.} \bibinfo{year}{2017}\natexlab{}.
\newblock \showarticletitle{Soccer video event detection using 3d convolutional
  networks and shot boundary detection via deep feature distance}. In
  \bibinfo{booktitle}{\emph{International Conference on Neural Information
  Processing}}. Springer, \bibinfo{pages}{440--449}.
\newblock


\bibitem[\protect\citeauthoryear{Mahaseni, Faizal, and Raj}{Mahaseni
  et~al\mbox{.}}{2021}]%
        {mahaseni2021spotting}
\bibfield{author}{\bibinfo{person}{Behzad Mahaseni}, \bibinfo{person}{Erma
  Rahayu~Mohd Faizal}, {and} \bibinfo{person}{Ram~Gopal Raj}.}
  \bibinfo{year}{2021}\natexlab{}.
\newblock \showarticletitle{Spotting Football Events Using Two-Stream
  Convolutional Neural Network and Dilated Recurrent Neural Network}.
\newblock \bibinfo{journal}{\emph{IEEE Access}}  \bibinfo{volume}{9}
  (\bibinfo{year}{2021}), \bibinfo{pages}{61929--61942}.
\newblock


\bibitem[\protect\citeauthoryear{Morra, Manigrasso, Canto, Gianfrate, Guarino,
  and Lamberti}{Morra et~al\mbox{.}}{2020b}]%
        {morra2020slicing}
\bibfield{author}{\bibinfo{person}{Lia Morra}, \bibinfo{person}{Francesco
  Manigrasso}, \bibinfo{person}{Giuseppe Canto}, \bibinfo{person}{Claudio
  Gianfrate}, \bibinfo{person}{Enrico Guarino}, {and} \bibinfo{person}{Fabrizio
  Lamberti}.} \bibinfo{year}{2020}\natexlab{b}.
\newblock \showarticletitle{Slicing and dicing soccer: Automatic detection of
  complex events from spatio-temporal data}. In
  \bibinfo{booktitle}{\emph{International Conference on Image Analysis and
  Recognition}}. Springer, \bibinfo{pages}{107--121}.
\newblock


\bibitem[\protect\citeauthoryear{Morra, Manigrasso, and Lamberti}{Morra
  et~al\mbox{.}}{2020a}]%
        {morra2020soccer}
\bibfield{author}{\bibinfo{person}{Lia Morra}, \bibinfo{person}{Francesco
  Manigrasso}, {and} \bibinfo{person}{Fabrizio Lamberti}.}
  \bibinfo{year}{2020}\natexlab{a}.
\newblock \showarticletitle{SoccER: Computer graphics meets sports analytics
  for soccer event recognition}.
\newblock \bibinfo{journal}{\emph{SoftwareX}}  \bibinfo{volume}{12}
  (\bibinfo{year}{2020}), \bibinfo{pages}{100612}.
\newblock


\bibitem[\protect\citeauthoryear{Nguyen, Liu, Prasad, and Han}{Nguyen
  et~al\mbox{.}}{2018}]%
        {nguyen2018weakly}
\bibfield{author}{\bibinfo{person}{Phuc Nguyen}, \bibinfo{person}{Ting Liu},
  \bibinfo{person}{Gautam Prasad}, {and} \bibinfo{person}{Bohyung Han}.}
  \bibinfo{year}{2018}\natexlab{}.
\newblock \showarticletitle{Weakly supervised action localization by sparse
  temporal pooling network}. In \bibinfo{booktitle}{\emph{Proceedings of the
  IEEE Conference on Computer Vision and Pattern Recognition}}.
  \bibinfo{pages}{6752--6761}.
\newblock


\bibitem[\protect\citeauthoryear{Pappalardo, Cintia, Rossi, Massucco,
  Ferragina, Pedreschi, and Giannotti}{Pappalardo et~al\mbox{.}}{2019}]%
        {pappalardo2019public}
\bibfield{author}{\bibinfo{person}{Luca Pappalardo}, \bibinfo{person}{Paolo
  Cintia}, \bibinfo{person}{Alessio Rossi}, \bibinfo{person}{Emanuele
  Massucco}, \bibinfo{person}{Paolo Ferragina}, \bibinfo{person}{Dino
  Pedreschi}, {and} \bibinfo{person}{Fosca Giannotti}.}
  \bibinfo{year}{2019}\natexlab{}.
\newblock \showarticletitle{A public data set of spatio-temporal match events
  in soccer competitions}.
\newblock \bibinfo{journal}{\emph{Scientific data}} \bibinfo{volume}{6},
  \bibinfo{number}{1} (\bibinfo{year}{2019}), \bibinfo{pages}{1--15}.
\newblock


\bibitem[\protect\citeauthoryear{Ramanathan, Huang, Abu-El-Haija, Gorban,
  Murphy, and Fei-Fei}{Ramanathan et~al\mbox{.}}{2016}]%
        {Ramanathan_2016_CVPR}
\bibfield{author}{\bibinfo{person}{Vignesh Ramanathan},
  \bibinfo{person}{Jonathan Huang}, \bibinfo{person}{Sami Abu-El-Haija},
  \bibinfo{person}{Alexander Gorban}, \bibinfo{person}{Kevin Murphy}, {and}
  \bibinfo{person}{Li Fei-Fei}.} \bibinfo{year}{2016}\natexlab{}.
\newblock \showarticletitle{Detecting Events and Key Actors in Multi-Person
  Videos}. In \bibinfo{booktitle}{\emph{Proceedings of the IEEE Conference on
  Computer Vision and Pattern Recognition (CVPR)}}.
\newblock


\bibitem[\protect\citeauthoryear{Read and Edwards}{Read and Edwards}{1997}]%
        {Read1997}
\bibfield{author}{\bibinfo{person}{Brenda Read} {and} \bibinfo{person}{Phyl
  Edwards}.} \bibinfo{year}{1997}\natexlab{}.
\newblock \bibinfo{booktitle}{\emph{Teaching Children to Play Games: A Resource
  for Primary Teachers}}.
\newblock \bibinfo{publisher}{{The English Sports Council}}.
\newblock
\showISBNx{978-1-86078-054-7}


\bibitem[\protect\citeauthoryear{Rein and Memmert}{Rein and Memmert}{2016}]%
        {rein2016big}
\bibfield{author}{\bibinfo{person}{Robert Rein} {and} \bibinfo{person}{Daniel
  Memmert}.} \bibinfo{year}{2016}\natexlab{}.
\newblock \showarticletitle{Big data and tactical analysis in elite soccer:
  future challenges and opportunities for sports science}.
\newblock \bibinfo{journal}{\emph{SpringerPlus}} \bibinfo{volume}{5},
  \bibinfo{number}{1} (\bibinfo{year}{2016}), \bibinfo{pages}{1--13}.
\newblock


\bibitem[\protect\citeauthoryear{Richly, Bothe, Rohloff, and Schwarz}{Richly
  et~al\mbox{.}}{2016}]%
        {richly2016recognizing}
\bibfield{author}{\bibinfo{person}{Keven Richly}, \bibinfo{person}{Max Bothe},
  \bibinfo{person}{Tobias Rohloff}, {and} \bibinfo{person}{Christian Schwarz}.}
  \bibinfo{year}{2016}\natexlab{}.
\newblock \showarticletitle{Recognizing compound events in spatio-temporal
  football data}. In \bibinfo{booktitle}{\emph{International Conference on
  Internet of Things and Big Data}}, Vol.~\bibinfo{volume}{2}. SCITEPRESS,
  \bibinfo{pages}{27--35}.
\newblock


\bibitem[\protect\citeauthoryear{Richly, Moritz, and Schwarz}{Richly
  et~al\mbox{.}}{2017}]%
        {richly2017utilizing}
\bibfield{author}{\bibinfo{person}{Keven Richly}, \bibinfo{person}{Florian
  Moritz}, {and} \bibinfo{person}{Christian Schwarz}.}
  \bibinfo{year}{2017}\natexlab{}.
\newblock \showarticletitle{Utilizing artificial neural networks to detect
  compound events in spatio-temporal soccer data}. In
  \bibinfo{booktitle}{\emph{Proceedings of the 2017 SIGKDD Workshop MiLeTS,
  Halifax, NS, Canada}}. \bibinfo{pages}{13--17}.
\newblock


\bibitem[\protect\citeauthoryear{Roth and Kr{\"o}ger}{Roth and
  Kr{\"o}ger}{2015}]%
        {Roth2015}
\bibfield{author}{\bibinfo{person}{Klaus Roth} {and} \bibinfo{person}{Christian
  Kr{\"o}ger}.} \bibinfo{year}{2015}\natexlab{}.
\newblock \bibinfo{booktitle}{\emph{{Ballschule: ein ABC f\"ur
  Spielanf\"anger}} (\bibinfo{edition}{fifth} ed.)}.
\newblock Number Band 1 in \bibinfo{series}{{Praxisideen : Schriftenreihe f\"ur
  Bewegung, Spiel und Sport Sportspiele}}. \bibinfo{publisher}{{Hofmann}},
  \bibinfo{address}{{Schorndorf}}.
\newblock
\showISBNx{978-3-7780-0015-1}


\bibitem[\protect\citeauthoryear{Sanford, Gorji, Hafemann, Pourbabaee, and
  Javan}{Sanford et~al\mbox{.}}{2020}]%
        {sanford2020group}
\bibfield{author}{\bibinfo{person}{Ryan Sanford}, \bibinfo{person}{Siavash
  Gorji}, \bibinfo{person}{Luiz~G Hafemann}, \bibinfo{person}{Bahareh
  Pourbabaee}, {and} \bibinfo{person}{Mehrsan Javan}.}
  \bibinfo{year}{2020}\natexlab{}.
\newblock \showarticletitle{Group Activity Detection from Trajectory and Video
  Data in Soccer}. In \bibinfo{booktitle}{\emph{Proceedings of the IEEE/CVF
  Conference on Computer Vision and Pattern Recognition Workshops}}.
  \bibinfo{pages}{898--899}.
\newblock


\bibitem[\protect\citeauthoryear{Sarkar, Chakrabarti, and
  Prasad~Mukherjee}{Sarkar et~al\mbox{.}}{2019}]%
        {sarkar2019generation}
\bibfield{author}{\bibinfo{person}{Saikat Sarkar}, \bibinfo{person}{Amlan
  Chakrabarti}, {and} \bibinfo{person}{Dipti Prasad~Mukherjee}.}
  \bibinfo{year}{2019}\natexlab{}.
\newblock \showarticletitle{Generation of ball possession statistics in soccer
  using minimum-cost flow network}. In \bibinfo{booktitle}{\emph{Proceedings of
  the IEEE/CVF Conference on Computer Vision and Pattern Recognition
  Workshops}}. \bibinfo{pages}{0--0}.
\newblock


\bibitem[\protect\citeauthoryear{Sigurdsson, Russakovsky, and Gupta}{Sigurdsson
  et~al\mbox{.}}{2017}]%
        {sigurdsson2017actions}
\bibfield{author}{\bibinfo{person}{Gunnar~A Sigurdsson}, \bibinfo{person}{Olga
  Russakovsky}, {and} \bibinfo{person}{Abhinav Gupta}.}
  \bibinfo{year}{2017}\natexlab{}.
\newblock \showarticletitle{What actions are needed for understanding human
  actions in videos?}. In \bibinfo{booktitle}{\emph{Proceedings of the IEEE
  international conference on computer vision}}. \bibinfo{pages}{2137--2146}.
\newblock


\bibitem[\protect\citeauthoryear{Sorano, Carrara, Cintia, Falchi, and
  Pappalardo}{Sorano et~al\mbox{.}}{2020}]%
        {sorano2020automatic}
\bibfield{author}{\bibinfo{person}{Danilo Sorano}, \bibinfo{person}{Fabio
  Carrara}, \bibinfo{person}{Paolo Cintia}, \bibinfo{person}{Fabrizio Falchi},
  {and} \bibinfo{person}{Luca Pappalardo}.} \bibinfo{year}{2020}\natexlab{}.
\newblock \showarticletitle{Automatic pass annotation from soccer VideoStreams
  based on object detection and LSTM}.
\newblock \bibinfo{journal}{\emph{arXiv preprint arXiv:2007.06475}}
  (\bibinfo{year}{2020}).
\newblock


\bibitem[\protect\citeauthoryear{Tomei, Baraldi, Calderara, Bronzin, and
  Cucchiara}{Tomei et~al\mbox{.}}{2021}]%
        {tomei2021rms}
\bibfield{author}{\bibinfo{person}{Matteo Tomei}, \bibinfo{person}{Lorenzo
  Baraldi}, \bibinfo{person}{Simone Calderara}, \bibinfo{person}{Simone
  Bronzin}, {and} \bibinfo{person}{Rita Cucchiara}.}
  \bibinfo{year}{2021}\natexlab{}.
\newblock \showarticletitle{Rms-net: Regression and masking for soccer event
  spotting}. In \bibinfo{booktitle}{\emph{2020 25th International Conference on
  Pattern Recognition (ICPR)}}. IEEE, \bibinfo{pages}{7699--7706}.
\newblock


\bibitem[\protect\citeauthoryear{Vanderplaetse and Dupont}{Vanderplaetse and
  Dupont}{2020}]%
        {vanderplaetse2020improved}
\bibfield{author}{\bibinfo{person}{Bastien Vanderplaetse} {and}
  \bibinfo{person}{Stephane Dupont}.} \bibinfo{year}{2020}\natexlab{}.
\newblock \showarticletitle{Improved soccer action spotting using both audio
  and video streams}. In \bibinfo{booktitle}{\emph{Proceedings of the IEEE/CVF
  Conference on Computer Vision and Pattern Recognition Workshops}}.
  \bibinfo{pages}{896--897}.
\newblock


\bibitem[\protect\citeauthoryear{Vats, Fani, Walters, Clausi, and Zelek}{Vats
  et~al\mbox{.}}{2020}]%
        {vats2020event}
\bibfield{author}{\bibinfo{person}{Kanav Vats}, \bibinfo{person}{Mehrnaz Fani},
  \bibinfo{person}{Pascale Walters}, \bibinfo{person}{David~A Clausi}, {and}
  \bibinfo{person}{John Zelek}.} \bibinfo{year}{2020}\natexlab{}.
\newblock \showarticletitle{Event detection in coarsely annotated sports videos
  via parallel multi-receptive field 1d convolutions}. In
  \bibinfo{booktitle}{\emph{Proceedings of the IEEE/CVF Conference on Computer
  Vision and Pattern Recognition Workshops}}. \bibinfo{pages}{882--883}.
\newblock


\bibitem[\protect\citeauthoryear{Wang, Girshick, Gupta, and He}{Wang
  et~al\mbox{.}}{2018}]%
        {NonLocal2018}
\bibfield{author}{\bibinfo{person}{Xiaolong Wang}, \bibinfo{person}{Ross
  Girshick}, \bibinfo{person}{Abhinav Gupta}, {and} \bibinfo{person}{Kaiming
  He}.} \bibinfo{year}{2018}\natexlab{}.
\newblock \showarticletitle{Non-local Neural Networks}.
\newblock \bibinfo{journal}{\emph{CVPR}} (\bibinfo{year}{2018}).
\newblock


\bibitem[\protect\citeauthoryear{Wittgenstein}{Wittgenstein}{1999}]%
        {Wittgenstein1999}
\bibfield{author}{\bibinfo{person}{Ludwig Wittgenstein}.}
  \bibinfo{year}{1999}\natexlab{}.
\newblock \bibinfo{booktitle}{\emph{Philosophical Investigations}
  (\bibinfo{edition}{second} ed.)}.
\newblock \bibinfo{publisher}{{Blackwell Publishers}},
  \bibinfo{address}{{Malden, Massachsetts}}.
\newblock


\bibitem[\protect\citeauthoryear{Xie, Wang, Liang, Deng, Cheng, Zhang, Chen,
  and Wu}{Xie et~al\mbox{.}}{2020}]%
        {xie2020passvizor}
\bibfield{author}{\bibinfo{person}{Xiao Xie}, \bibinfo{person}{Jiachen Wang},
  \bibinfo{person}{Hongye Liang}, \bibinfo{person}{Dazhen Deng},
  \bibinfo{person}{Shoubin Cheng}, \bibinfo{person}{Hui Zhang},
  \bibinfo{person}{Wei Chen}, {and} \bibinfo{person}{Yingcai Wu}.}
  \bibinfo{year}{2020}\natexlab{}.
\newblock \showarticletitle{PassVizor: Toward better understanding of the
  dynamics of soccer passes}.
\newblock \bibinfo{journal}{\emph{IEEE Transactions on Visualization and
  Computer Graphics}} (\bibinfo{year}{2020}).
\newblock


\bibitem[\protect\citeauthoryear{Yu, Lei, and Hu}{Yu et~al\mbox{.}}{2019}]%
        {yu2019soccer}
\bibfield{author}{\bibinfo{person}{Junqing Yu}, \bibinfo{person}{Aiping Lei},
  {and} \bibinfo{person}{Yangliu Hu}.} \bibinfo{year}{2019}\natexlab{}.
\newblock \showarticletitle{Soccer video event detection based on deep
  learning}. In \bibinfo{booktitle}{\emph{International Conference on
  Multimedia Modeling}}. Springer, \bibinfo{pages}{377--389}.
\newblock


\end{thebibliography}

%\newpage
% \input{appendix}

\end{document}